\documentclass[sn-mathphys]{sn-jnl}% Math and Physical Sciences Reference Style
\jyear{2021}%

\theoremstyle{thmstyleone}%
\theoremstyle{thmstyletwo}%

\theoremstyle{thmstylethree}%

\graphicspath{{fig/}}

\usepackage[utf8]{inputenc}
\usepackage{comment}
\usepackage{commath} % for \abs{}
\usepackage{csquotes}
\usepackage{hhline}
\usepackage{tablefootnote}
\usepackage{tabularx}
\usepackage{xcolor}
\usepackage{xspace}
\usepackage{multirow}

\newcommand{\FEVER}{\textsc{FEVER}\xspace}
\newcommand{\FEVERNLI}{\textsc{FEVER-NLI}\xspace}
\newcommand{\Wikipedia}{\textsc{Wikipedia}\xspace}
\newcommand{\MediaWiki}{\textsc{MediaWiki}\xspace}
\newcommand{\FCZ}{\textsc{CsFEVER}\xspace}
\newcommand{\FCZNLI}{\textsc{CsFEVER-NLI}\xspace}
\newcommand{\FEN}{\textsc{EnFEVER}\xspace}
\newcommand{\FDAN}{\textsc{DanFEVER}\xspace}
\newcommand{\CTK}{\textsc{CTKFacts}\xspace}
\newcommand{\CTKNLI}{\textsc{CTKFactsNLI}\xspace}
\newcommand{\Anserini}{\textsc{Anserini}\xspace}
\newcommand{\DrQA}{\textsc{DrQA}\xspace}
\newcommand{\BERT}{\textsc{Bert}\xspace}
\newcommand{\RoBERTa}{\textsc{RoBERTa}\xspace}
\newcommand{\MBERT}{\textsc{M-Bert}\xspace}
\newcommand{\SMBERT}{\textsc{Sentence M-Bert}\xspace}
\newcommand{\ColBERT}{\textsc{ColBert}\xspace}
\newcommand{\SlavicBERT}{\textsc{SlavicBERT}\xspace}
\newcommand{\CZERT}{\textsc{Czert}\xspace}
\newcommand{\RobeCzech}{\textsc{RobeCzech}\xspace}
\newcommand{\XLM}{\textsc{XLM-RoBERTa}\xspace}
\newcommand{\FERNETC}{\textsc{FERNET-C5}\xspace}
\newcommand{\FERNETN}{\textsc{FERNET-News}\xspace}
\newcommand{\XLMSQUAD}{\textsc{XLM-RoBERTa @ SQuAD2}\xspace}
\newcommand{\XLMXNLI}{\textsc{XLM-RoBERTa @ XNLI}\xspace}

\newcommand{\train}{\textsf{train}\xspace}
\newcommand{\dev}{\textsf{dev}\xspace}
\newcommand{\test}{\textsf{test}\xspace}

\newcommand{\SUP}{\texttt{SUPPORTS}}
\newcommand{\REF}{\texttt{REFUTES}}
\newcommand{\NEI}{\texttt{NEI}}

\newcommand{\Tzero}{{$\textsf{T}_{\textsf{0}}$}}
\newcommand{\Tone}{{$\textsf{T}_{\textsf{1}}$}}
\newcommand{\ToneA}{{$\textsf{T}_{\textsf{1a}}$}}
\newcommand{\ToneB}{{$\textsf{T}_{\textsf{1b}}$}}
\newcommand{\Ttwo}{{$\textsf{T}_{\textsf{2}}$}}

\newcommand{\q}[1]{``#1''}
\newcommand{\qit}[1]{\textit{``#1''}}
\newcommand{\revision}[1]{{\color{black}#1}}
\newcommand{\jd}[1]{{\color{orange}\textbf{jd: }#1}}

\newcommand{\todo}[1]{{\color{red}\colorbox{yellow}{\textbf{TODO: }}#1}}

\definecolor{delim}{RGB}{20,105,176}
\colorlet{punct}{red!60!black}
\colorlet{numb}{magenta!60!black}

\lstdefinelanguage{json}{
    basicstyle=\normalfont\ttfamily,
    numbers=left,
    numberstyle=\scriptsize,
    stepnumber=1,
    numbersep=8pt,
    showstringspaces=false,
    breaklines=true,
    frame=lines,
    backgroundcolor=\color{white},
    literate=
     *{0}{{{\color{numb}0}}}{1}
      {1}{{{\color{numb}1}}}{1}
      {2}{{{\color{numb}2}}}{1}
      {3}{{{\color{numb}3}}}{1}
      {4}{{{\color{numb}4}}}{1}
      {5}{{{\color{numb}5}}}{1}
      {6}{{{\color{numb}6}}}{1}
      {7}{{{\color{numb}7}}}{1}
      {8}{{{\color{numb}8}}}{1}
      {9}{{{\color{numb}9}}}{1}
      {:}{{{\color{punct}{:}}}}{1}
      {,}{{{\color{punct}{,}}}}{1}
      {\{}{{{\color{delim}{\{}}}}{1}
      {\}}{{{\color{delim}{\}}}}}{1}
      {[}{{{\color{delim}{[}}}}{1}
      {]}{{{\color{delim}{]}}}}{1},
}

\usepackage{enumitem}
\SetLabelAlign{myright}{\hss\llap{$#1$}}
\newlist{where}{description}{1}
\setlist[where]{labelwidth=2cm, labelsep=1em, leftmargin=!, align=myright, font=\normalfont}

\begin{document}

% \title[\CTK: fact-checking dataset based on Czech news corpus]{\CTK: fact-checking dataset based on Czech news corpus}
\begin{comment}
  TODO:
  (2.) The second dataset, CTKFacts, is not available for publication for
licensing reasons. A variant to circumvent the problem is released for
a different task, the NLI.

    Based on the title, one would expect two Czech datasets for fact-checking, but
in the end, one gets none. What we actually do get, are two Czech datasets for
NLI. Besides the two NLI datasets, the described method for creating
a fact-checking dataset from news articles (Section 4) is interesting. The
submission is still worthwhile for presentation and after some revisions, it has
the potential to be a fine submission, but I suggest that it should
be renamed and reframed accordingly to fit its actual main contributions, which
are:  

- two Czech NLI datasets,
- the process of annotation and tools for creating a fact-checking dataset from
 news (Section 4).
\end{comment}

\newcommand{\papertitle}{\FCZ and \CTK: Acquiring Czech data for fact verification}
\title[\papertitle]{\revision{\papertitle}}


\author*[1]{\revision{\fnm{Herbert} \sur{Ullrich}}}\email{ullriher@fel.cvut.cz}
\equalcont{These authors contributed equally to this work.}

\author*[1]{\revision{\fnm{Jan} \sur{Drchal}}}\email{drchajan@fel.cvut.cz}
\equalcont{These authors contributed equally to this work.}

\author[1]{\fnm{Martin} \sur{R\'{y}par}}\email{ryparmar@fel.cvut.cz}

\author[2]{\fnm{Hana} \sur{Vincourov\'{a}}}\email{77015105@fsv.cuni.cz}

\author[2]{\fnm{V\'{a}clav} \sur{Moravec}}\email{vaclav.moravec@fsv.cuni.cz}

\affil*[1]{\orgdiv{Artificial Intelligence Center}, \orgname{Faculty of Electrical Engineering, Czech Technical University in Prague}, \orgaddress{\street{Charles Square 13}, \city{Prague~2}, \postcode{120 00}, \country{Czech Republic}}}
\affil[2]{\orgdiv{Department of Journalism}, \orgname{Faculty of Social Sciences, Charles University}, \orgaddress{\street{Smetanovo n\'{a}b\v{r}e\v{z}\'{i} 6}, \city{Prague~1}, \postcode{110 01},  \country{Czech Republic}}}

\begin{comment}
  In this paper, we present \revision{freshly collected data} for automated fact-checking \revision{in Czech language}, which is a task commonly modeled as a classification of textual claim veracity w.r.t. a corpus of trusted ground truths.
We consider 3 classes: \SUP{}, \REF{} complemented with evidence documents or \NEI{} (Not Enough Info) alone. 
Our first dataset, \FCZ, has 127,328 claims.
It is an automatically generated Czech version of the large-scale \FEVER dataset built on top of \Wikipedia corpus.
We take a hybrid approach of machine translation and document alignment; the approach, and the tools we provide, can be easily applied to other languages.
The second dataset, \CTK of 3,097 claims, is annotated using the corpus of 2.2M articles of Czech News Agency.
We present its extended annotation methodology based on the \FEVER approach.
% Most notably, we describe a method to automatically generate wider claim contexts (dictionaries) for non-hyperlinked corpora.
We analyze both datasets for spurious cues -- annotation patterns leading to model overfitting.
\CTK is further examined for inter-annotator agreement, thoroughly cleaned, and a typology of common annotator errors is extracted.
Finally, we provide baseline models for all stages of the fact-checking pipeline.
\end{comment}

\abstract{
\revision{In this paper, we examine several methods of acquiring Czech data for automated fact-checking, which is a task commonly modeled as a classification of textual claim veracity w.r.t. a corpus of trusted ground truths.
We attempt to collect sets of data in form of a factual claim, evidence within the ground truth corpus, and its veracity label (\textit{supported}, \textit{refuted} or \textit{not enough info}).
As a first attempt, we generate a Czech version of the large-scale \FEVER dataset built on top of \Wikipedia corpus.
We take a hybrid approach of machine translation and document alignment; the approach and the tools we provide can be easily applied to other languages.
We discuss its weaknesses and inaccuracies, propose a future approach for their cleaning and publish the 127k resulting translations, as well as a version of such dataset reliably applicable for the Natural Language Inference task - the \FCZNLI.
Furthermore, we collect a novel dataset of 3,097 claims, which is annotated using the corpus of 2.2M articles of Czech News Agency.
We present its extended annotation methodology based on the \FEVER approach, and, as the underlying corpus is kept a trade secret, we also publish a standalone version of the dataset for the task of Natural Language Inference we call \CTKNLI.
% Most notably, we describe a method to automatically generate wider claim contexts (dictionaries) for non-hyperlinked corpora.
We analyze both acquired datasets for spurious cues -- annotation patterns leading to model overfitting.
\CTK is further examined for inter-annotator agreement, thoroughly cleaned, and a typology of common annotator errors is extracted.
Finally, we provide baseline models for all stages of the fact-checking pipeline and publish the NLI datasets, as well as our annotation platform and other experimental data.\footnote{\url{https://github.com/aic-factcheck/csfever-and-ctkfacts-paper}.}}
}

% max 6 Keywords



\maketitle

% \input{src/revision_note}
%!TEX ROOT=../lrec2021.tex

\begin{comment}
TOFINALIZE: 
  2. Pragmatic (extra linguistic) knowledge and context is crucial. For
    example, imagine an example: "Country X did not invade country Y", Z said.
    Here the statement depends on the Z's position. Of course this is more
    a philosophical remark and I do not see how you can prevent anyone of
    misusing the approach, but maybe you could consider adding one or two
    sentences if you also agree this should be commented upon?
\end{comment}
\section{Introduction}\label{sec:intro}

In the current highly connected online society, the ever-growing information influx eases the spread of false or misleading news.
The omnipresence of fake news motivated formation of fact-checking organizations such as AFP Fact Check,\footnote{\url{https://factcheck.afp.com/}} International Fact-Checking Network,\footnote{\url{https://www.poynter.org/ifcn/}} PolitiFact,\footnote{\url{https://www.politifact.com/}} Poynter,\footnote{\url{https://www.poynter.org/}} Snopes,\footnote{\url{https://www.snopes.com/}} and many others.
At the same time, many tools for fake news detection and fact-checking are being developed: ClaimBuster~\cite{hassan2017claimbuster}, ClaimReview\footnote{\url{https://www.claimreviewproject.com/}} or CrowdTangle;\footnote{\url{https://www.crowdtangle.com/}} see~\cite{zeng2021fcsurvey} for more examples.
Many of these are based on machine learning technologies aimed at image recognition, speech to text, or Natural Language Processing (NLP).
This \revision{article} deals with the latter, focusing on automated fact-checking (hereinafter also referred to as \textit{fact verification}).

Automated fact verification is a complex NLP task~\cite{thorne2018automated} in which the veracity of a textual \textit{claim} gets evaluated with respect to a ground truth corpus.
The output of a fact-checking system gives a classification of the claim -- conventionally varying between \textit{supported}, \textit{refuted} and \textit{not enough information} available in corpus. 
For the \textit{supported} and \textit{refuted} outcomes it further supplies the \textit{evidence}, i.e., a list of documents that explain the verdict.
Fact-checking systems typically work in two stages~\cite{fever2018}. 
In the first stage, based on the input \textit{claim}, the Document Retrieval (DR) module selects the \textit{evidence}.
In the second stage, the Natural Language Inference (NLI) module matches the \textit{evidence} with the \textit{claim} and provides the final verdict.    
Table~\ref{tab:fc_example} shows an example of data used to train the fact-checking systems of this type.

\begin{table}
  \begin{center}
  \begin{minipage}{0.83\textwidth}
  \caption{Truncated example from \CTK \train set.}\label{tab:fc_example}
  \begin{tabular}{p{\linewidth}}
  \toprule
  \textbf{Claim:} Spojené státy americké hraničí s Mexikem.\\ 
  \textbf{EN Translation:} \textit{The United States of America share borders with Mexico.}\\ 
  \midrule
  \textbf{Verdict:} \SUP\\ 
  \midrule
  \textbf{Evidence 1:} \q{Mexiko a USA sdílejí 3000 kilometrů dlouhou hranici, kterou ročně překročí tisíce Mexičanů v naději na lepší životní podmínky (\dots)}\\ 
   \textbf{EN:} \textit{Mexico and the U.S. share a 3,000-kilometre border, thousands of Mexicans cross each year in hopes of better living conditions (\dots)}\\\\
  \textbf{Evidence 2:} \q{Mexiko také nelibě nese, že Spojené státy stále budují na vzájemné, několik tisíc kilometrů dlouhé hranici zeď, která má zabránit fyzickému ilegálnímu přechodu Mexičanů do USA (\dots)}\\ 
   \textbf{EN:} \textit{Mexico is also uncomfortable with the fact that the United States is still building a wall on their mutual, several thousand-mile borders to prevent Mexicans from physically crossing illegally into the U.S. (\dots)}\\
  \botrule
  \end{tabular}
  \end{minipage}
  \end{center}
\end{table}

Current state-of-the-art methods applied to the domain of automated fact-checking are typically based on large-scale neural language models~\cite{fever2018b}, which are notoriously data-hungry. 
\revision{While there is a reasonable number of quality datasets available for high-profile world languages~\cite{zeng2021fcsurvey}, the situation for the most other languages is significantly less favorable.}
Also, most available large-scale datasets are built on top of \Wikipedia~\cite{fever2018,aly2021feverous,schuster-etal-2021-vitaminc,sathe2020automated}. 
While encyclopedic corpora are convenient for dataset annotation, these are hardly the only eligible sources of the ground truth. 

We argue that corpora of verified news articles used as claim verification datasets are a relevant alternative to encyclopedic corpora.
Advantages are clear: the amount and detail of information covered by news reports are typically higher.
Furthermore, the news articles typically inform on recent events attracting public attention, which also inspire new fake or misleading claims spreading throughout the online space.

On the other hand, news articles address a more varied range of issues and have a more complex structure from the NLP perspective.
While encyclopedic texts are typically concise and focused on facts, the style of news articles can vary wildly between different documents or even within a single article.
For example, it is common that a report-style article is intertwined with quotations and informative summaries.
Also, claim validity might be obscured by complex temporal or personal relationships: a past quotation like \qit{Janet Reno will become a member of the Cabinet.} may or may not support the claim \qit{Janet Reno was the member of the Cabinet.}\footnote{\revision{And the veracity of such claim may be further nuanced by the affiliation and bias of the speaker.}}
This depends on, firstly, which date we verify the claim validity to, and secondly, who was or what was the competence of the quotation's author.
Note that similar problems are less likely in encyclopedia-based datasets like \FEVER\cite{fever2018}.

The contributions of this paper are as follows:
\begin{enumerate} % TODO: Better rewrite the csfever part
    \item \textbf{\revision{\FEVER localization scheme (and \FCZ case study)}:} We propose an experimental localization \revision{scheme} of the large-scale \FEVER~\cite{fever2018} fact-checking dataset, utilizing the public \MediaWiki interlingual document alignment of \Wikipedia articles and a MT-based claim transduction.
    We publish our procedure to be used for other languages, and analyze its pitfalls\revision{. We observe the types of translation loss on our \FCZ dataset obtained through this procedure and approximate their frequency}.
    We denote the original English \FEVER as \FEN in the following sections to distinguish various language mutations.
    \item \textbf{CTKFacts:} we introduce a new Czech fact-checking dataset manually annotated on top of approximately two million Czech News Agency\footnote{\url{https://www.ctk.eu/}} news reports from 2000--2020.
    Inspired by \FEVER, we provide an updated and extended annotation methodology aimed at annotations of news corpora, and we also make available an open-source annotation platform.
    The \textit{claim generation} as well as \textit{claim labeling} is centered around limited  knowledge context (denoted \textit{dictionary} in~\cite{fever2018}), which is trivial to construct for hyperlinked textual corpora such as \Wikipedia.
    We present a novel approach based on document retrieval and clustering.
    The method automatically generates dictionaries, which are composed of both relevant and semantically diverse documents, and does not depend on any inter-document linking.
    \revision{We present a manually cleaned set of 3k labeled claims from our annotations with 63\% Fleiss' $\kappa$-agreement, backed by evidence from the proprietary CTK corpus. We also publish its standalone (evidence-included) version we call \CTKNLI{}.}
    \item We provide a detailed analysis of the \CTK dataset\revision{, including the empirical (based on manual conflict resolution)}, and \textit{spurious cue analysis}, where the latter detects annotation patterns possibly leading to overfitting of the NLP models.
    For comparison, we analyze the spurious cues of \FCZ as well.
    We construct an annotation cleaning scheme that involves both manual and semi-automated procedures, and we use it to refine the final version of the \CTK dataset.
    We also provide classification and discussion of common annotation errors for future improvements of the annotation methodology.
    \item We present baseline models for both DR and NLI stages as well as for the full fact-checking pipeline.
    \item We publicly release the \CTK dataset as well as the experimental \FCZ data, used source code and the baseline models. \revision{Due to the weaknesses of the \FCZ{} dataset revealed in Section~\ref{sec:fevercs} and to the licensing of the ground-truth corpus underlying \CTK, we also publish their NLI versions \FCZNLI and \CTKNLI that can be used on their own and do not suffer from the transduction noise.} 
    Data, tools, and models are available under the \textsf{CC BY-SA 3.0} license.
\end{enumerate}

This article is structured as follows: in Section~\ref{sec:related_work}, we give an overview of the related work.
Section~\ref{sec:fevercs} describes our experimental method to localize the \FEN dataset using the \MediaWiki alignment.
We generate the Czech language \FCZ dataset with it and analyze its validity.
In~Section~\ref{sec:ctkfacts}, we introduce the novel \CTK dataset.
We describe its annotation methodology, data cleaning, and postprocessing, as well as analysis of the inter-annotator agreement.
Section~\ref{sec:analysis} analyzes spurious cues for both \FCZ and \CTK.
In Section~\ref{sec:baseline}, we present the baseline models.
Section~\ref{sec:conclusion} concludes with an overall discussion of the results and with remarks for future research.

% \begin{figure}[H]
% \begin{lstlisting}[language=json]
% {
%   "id": 36242,
%   "verifiable": "VERIFIABLE",
%   "label": "REFUTES",
%   "claim": "Mud was made before Matthew McConaughey was born.",
%   "evidence": [
%     [
%       [52443, 62408, "Mud_-LRB-2012_film-RRB-", 1],
%       [52443, 62408, "Matthew_McConaughey", 0]
%     ],
%     [
%       [52443, 62409, "Mud_-LRB-2012_film-RRB-", 0]
%     ]
%   ]
% }
% \end{lstlisting}
%     \caption{Example \FEN \texttt{REFUTES} annotation with two possible evidence sets. \todo{Take example from \FCZ.}}
%     \label{list:fever}
% \end{figure}
%!TEX ROOT=../lrec2021.tex
\section{Related work}\label{sec:related_work}
This section describes datasets and models related to the task of automated fact-checking of textual claims.
More general overview of the state-of-the-art can be found in~\cite{zeng2021fcsurvey} or~\cite{murayama2021dataset}.

Emergent~\cite{ferreira2016emergent} dataset is based on news; it contains 300 claims and 2k+ articles, however, it is limited to headlines.
Due to the dataset size, only simple models classifying to three classes (\textit{for}, \textit{against}, and \textit{observing}) are presented.
Described models are fed BoW vectors and feature-engineered attributes.

Wang in~\cite{wang2017liar} presents another dataset of 12k+ claims, working with 5 classes (\textit{pants-fire}, \textit{false}, \textit{barely-true}, \textit{half-true}, \textit{mostly-true}, and \textit{true}).
Each verdict includes a justification. 
However, evidence sources are missing.
The models presented in the paper are claim-only, i.e., they deal with surface-level linguistic cues only.
The author further experiments with speaker-related meta-data.

Fact Extraction and VERification (\FEVER)~\cite{fever2018} is a large dataset of 185k+ claims covering the overall fact-checking pipeline. 
It is based on abstracts of 50k most visited pages of English \Wikipedia.
Authors present complex annotation methodology that involves two stages: the \textit{claim generation} in which annotators firstly create a true \textit{initial claim} supported by a random \Wikipedia source article with context extended by the \textit{dictionary} constructed from pages linked from the source article.
The \textit{initial claim} is further \textit{mutated} by rephrasing, negating and other operations.
The task of the second \textit{claim labeling} stage is to provide the \textit{evidence} as well as give the final verdict: \texttt{SUPPORTS}, \texttt{REFUTES} or \texttt{NEI}, where the latter stands for the \revision{\q{\textit{not enough information}}} label.
Fact Extraction and VERification Over Unstructured and Structured information (\textsc{FEVEROUS})~\cite{aly2021feverous} adds 87k+ claims including evidence based on \Wikipedia table cells.
The size of \FEVER data facilitates modern deep learning NLP methods.
The \FEVER authors host annual workshops involving competitions, with results described in~\cite{fever2018b} and~\cite{thorne2019fever2}.

\textsc{MultiFC}~\cite{augenstein2019multifc} is a 34k+ claim dataset sourcing its claims from 26 fact checking sites.
The evidence documents are retrieved via Google Search API as the ten highest-ranking results.
This approach significantly deviates from the \FEVER-like datasets as the ground-truth is not limited by a closed-world corpus, which limits the trustworthiness of the retrieved evidence.
Also, similar data cannot be utilized to train the DR models.

\textsc{WikiFactCheck-English}~\cite{sathe2020automated} is another recent \Wikipedia-based large dataset of 124k+ claims and further 34k+ ones including claims refuted by the same evidence.
The claims are accompanied by \textit{context}.
The evidence is based on \Wikipedia articles as well as on the linked documents. 

Considering other than English fact-checking datasets, the situation is less favorable.
Recently, Gupta et al.~\cite{gupta2021xfact} released a multilingual (25 languages) dataset of 31k+ claims annotated by seven veracity classes.
Similarly to the \textsc{MultiFC}, evidence is retrieved via Google Search API.
The experiments with the multilingual \BERT~\cite{devlin2019bert} model show that the gain from including the evidence is rather limited when compared to claim-only models.
FakeCovid~\cite{shahi2020fakecovid} is a multilingual (40 languages) dataset of 5k+ news articles.
The dataset focuses strictly on the COVID-19 topic.
Also, it does not supply evidence in a raw form -- human fact-checker argumentation is provided instead. 
Kazemi et al.~\cite{kazemi2021claim} released two multilingual (5 languages) datasets, these are, however, aimed at \textit{claim detection} (5k+ examples) and \textit{claim matching} (2k+ claim pairs).

In the Czech locale, the most significant machine-learnable dataset is the \textsc{Demagog} dataset~\cite{priban-etal-2019-machine} based on the fact-checks of the Demagog\footnote{\url{https://demagog.cz/}} organisation.
The dataset contains 9k+ claims in Czech (and 15k+ in Slovak and Polish) labeled with a veracity verdict and speaker-related metadata, such as name and political affiliation.
The verdict justification is given in natural language, often providing links from social networks, government-operated webpages, etc.
While the metadata is appropriate for statistical analyses, the justification does not come from a closed knowledge base that could be used in an automated scheme.

The work most related to ours was presented by the authors of~\citep{binau2020danish,norregaard2021danfever}, who published a Danish version of \FEN called \FDAN.
Unlike our \FCZ dataset, \FDAN was annotated by humans.
Given the limited number of annotators, it includes significantly fewer claims than \FEN (6k+ as opposed to 185k+).
% , which makes it less appealing for the state-of-the-art neural models.

%!TEX ROOT=../lrec2021.tex

\section{\FCZ}\label{sec:fevercs}
In this section, we introduce a developmental \FCZ dataset intended as a Czech localization of the large-scale English \FEN dataset.
It consists of claims and veracity labels justified with pointers to data within the Czech \Wikipedia dump.

A \revision{typical} approach to automatically build such a dataset from the \FEN data would be to employ machine translation (MT) methods for both claims and \Wikipedia articles. 
\begin{comment} TOFINALIZE:
    In Section 3, I humbly disagree with the adjective "straightforward" in "A
straightforward approach..." I think machine translation or manual translation
(in smaller datasets) is not only the straightforward approach, but (generally)
typical, standard and would be optimal even in this case, only if it scaled up.
Because it does not scale up, you were forced to look for other options, which
in result brings up some problems (difficult or impossible sentence-to-sentence
alignment). So I would suggest refraining from the adjective "straightforward",
which IMHO has the misleading slight flavor of "brute force but suboptimal",
while at the same does not convey "optimal, but not scalable". This comment is
only a matter of reading between the lines and may be only my idiosyncratic
view.
\end{comment}
While MT methods are recently reaching maturity~\cite{dabre2020mtsurvey,popel2020Transforming}, the problem lies in the high computational complexity of such translation.
While using the state-of-the-art MT methods to translate the claims (2.2M words) is a feasible way of acquiring data, the translation of all \Wikipedia articles is a much costlier task, as only their abstracts have a total of 513M words \revision{corresponding to 452k pages} (measuring the June 2017 dump used in~\cite{fever2018}).

However, in NLP research, \Wikipedia localizations are often considered a \textit{comparable corpus}~\cite{vstromajerova2016between,mohammadi2010parallelwiki,chu-etal-2014-constructing,Fiser2015poornet,Althobaiti2021Wikiparallel}, that is, a corpus of texts that share a~domain and properties.
Furthermore, partial alignment is often revealed between \Wikipedia locales, either on the level of article titles~\cite{Fiser2015poornet}, or specific sentences~\cite{vstromajerova2016between} -- much like in \textit{parallel} corpora. 
We hypothesize there may be a sufficient document-level alignment between Czech and English \Wikipedia abstracts that were used to annotate the \FEN dataset, as in both languages the abstracts are used to summarize basic facts about the same real-world entity.

In order to validate this hypothesis, and to obtain experimental large-scale data for our task, we proceed to localize the \FEN dataset using such an alignment derived from the \Wikipedia interlanguage linking available on \MediaWiki.\footnote{\url{https://www.mediawiki.org/}}
In the following sections, we discuss the output quality and information loss, and we outline possible uses of the resulting dataset.

\subsection{Method}\label{fcz-method}
Our approach to generating \FCZ from the openly available \FEN dataset can be summarized by the following steps:
\begin{enumerate}
    \item Fix a version of Wikipedia dump in the target language to be the verified corpus. 

    \item Map each \Wikipedia article referred in the evidence sets to a corresponding localized article using \MediaWiki \textsc{API}.\footnote{\url{https://www.mediawiki.org/wiki/API:REST_API}}
    If no localization is available for an article, remove all \textit{evidence sets} in which it occurs.

    \item Remove all \SUP{} and \REF{} data points having empty evidence.

    \item Apply MT method of choice to all claims.

    \item Re-split the dataset to \train, \dev, and \test so that the \dev and \test veracity labels are balanced.
\end{enumerate}

Before we explore the data, let us discuss the caveats of the scheme itself.
Firstly, the evidence sets are not guaranteed to be exhaustive -- no human annotations in the target language were made to detect whether there are new ways of verifying the claims using the target version of \Wikipedia (in fact, this does not hold for \FEN either, as its evidence-recall was estimated to be 72.36\%~\cite{fever2018}).

Secondly, even if our document-alignment hypothesis is valid on the level of abstracts, sentence-level alignment is not guaranteed.
Its absence invalidates the \FEN evidence format, where evidence is an array of \Wikipedia \textit{sentence} identifiers.
The problem could, however, be addressed by altering the evidence \textit{granularity} of the dataset, i.e., using whole documents to prove or refute the claim, rather than sentences.
Recent research on long-input processing language models~\cite{beltagy2020longformer,kitaev2020reformer,xiong2021nystromformer} is likely to make this simplification less significant. 

\revision{Lastly, the step 5. might allow a slight leakage of information between the splits -- while it is guaranteed that no claim appears in two splits simultaneously, two claims extracted from the same \Wikipedia article may -- however, such claims are typically its different \textit{mutations}, independent of each other's veracity.
This can not be fixed using publicly available data, if we're aiming for balanced \dev{}, \test{} splits.}
\subsection{Results}
Following our scheme from Section~\ref{fcz-method}, we used the June 2020 Czech \Wikipedia dump parsed into a database of plain text articles using the \texttt{wikiextractor}\footnote{\url{https://github.com/attardi/wikiextractor}} package and only kept their abstracts.

In order to translate the claims, we have empirically tested three available state-of-the-art English--Czech machine translation engines (data not shown here).
Namely, these were: Google Cloud Translation API,\footnote{\url{https://cloud.google.com/translate}} \textsc{CUBBITT}~\cite{popel2020Transforming} and \textsc{DeepL}.\footnote{\url{https://deepl.com/}} As of March $17^{th}$ 2021 we observed \textsc{DeepL} to give the best results.
Most importantly, it turned out to be robust w.r.t. homographs and faithful to the conventional translation of named entities (such as movie titles, which are very common amongst the 50k most popular \Wikipedia articles used in~\cite{fever2018}).

Finally, during the localization process, we have been able to locate Czech versions of 6,578 out of 12,633 \Wikipedia articles needed to infer the veracity of all \FEN claims.
Omitting the evidence sets that are not fully contained by the Czech \Wikipedia and omitting \SUP/\REF{} claims with empty evidence, we arrive to 127,328 claims that can hypothetically be fully (dis-)proven in at least one way using the Czech \Wikipedia abstracts corpus, which is 69\% of the total 185,445 \FEN claims.

We release the resulting dataset publicly in the HuggingFace datasets repository.\footnote{\url{https://huggingface.co/datasets/ctu-aic/csfever}}
In Table~\ref{tab:fevercs-overview} we show the dataset class distribution.
It is roughly proportional to that of \FEN.
Similarly to~\cite{fever2018}, we have opted for label-balanced \dev and \test splits, in order to ease evaluation of biased predictors.

\begin{table}
    \begin{center}
        \begin{minipage}{\textwidth}
            \caption{Label distribution in \FCZ dataset as oposed to the \FEN. The \test split of \FEN is not public.}\label{tab:fevercs-overview}
            \begin{tabular*}{\textwidth}{@{\extracolsep{\fill}}lllllll@{\extracolsep{\fill}}}
                \toprule
                & \multicolumn{3}{@{}c@{}}{\FCZ} & \multicolumn{3}{@{}c@{}}{\FEN}\\
                \cmidrule{2-4}\cmidrule{5-7}
                Split & {\texttt{SUPPORTS}} & \texttt{REFUTES}  & \texttt{NEI} & {\texttt{SUPPORTS}} & \texttt{REFUTES}  & \texttt{NEI}\\
                \midrule
                \train & 53,542 & 18,149 & 35,639 & 80,035 &29,775& 35,639\\
                \dev & 3,333 & 3,333 & 3,333 & 6,666 & 6,666 & 6,666\\
                \test & 3,333 & 3,333 & 3,333& 6,666 & 6,666 & 6,666\\
                \botrule
            \end{tabular*}
        \end{minipage}
    \end{center}
\end{table}

\subsubsection{Validity}
\label{fcz-validity}

In order to validate our hypothesis that the Czech \Wikipedia abstracts \revision{support and refute the same claims as} their English counterparts, we have sampled 1\% (1257) verifiable claim-evidence pairs from the \FCZ dataset and annotated their validity.

\begin{figure}[H]
    \makebox[\textwidth][c]{\includegraphics[width=0.6\textwidth]{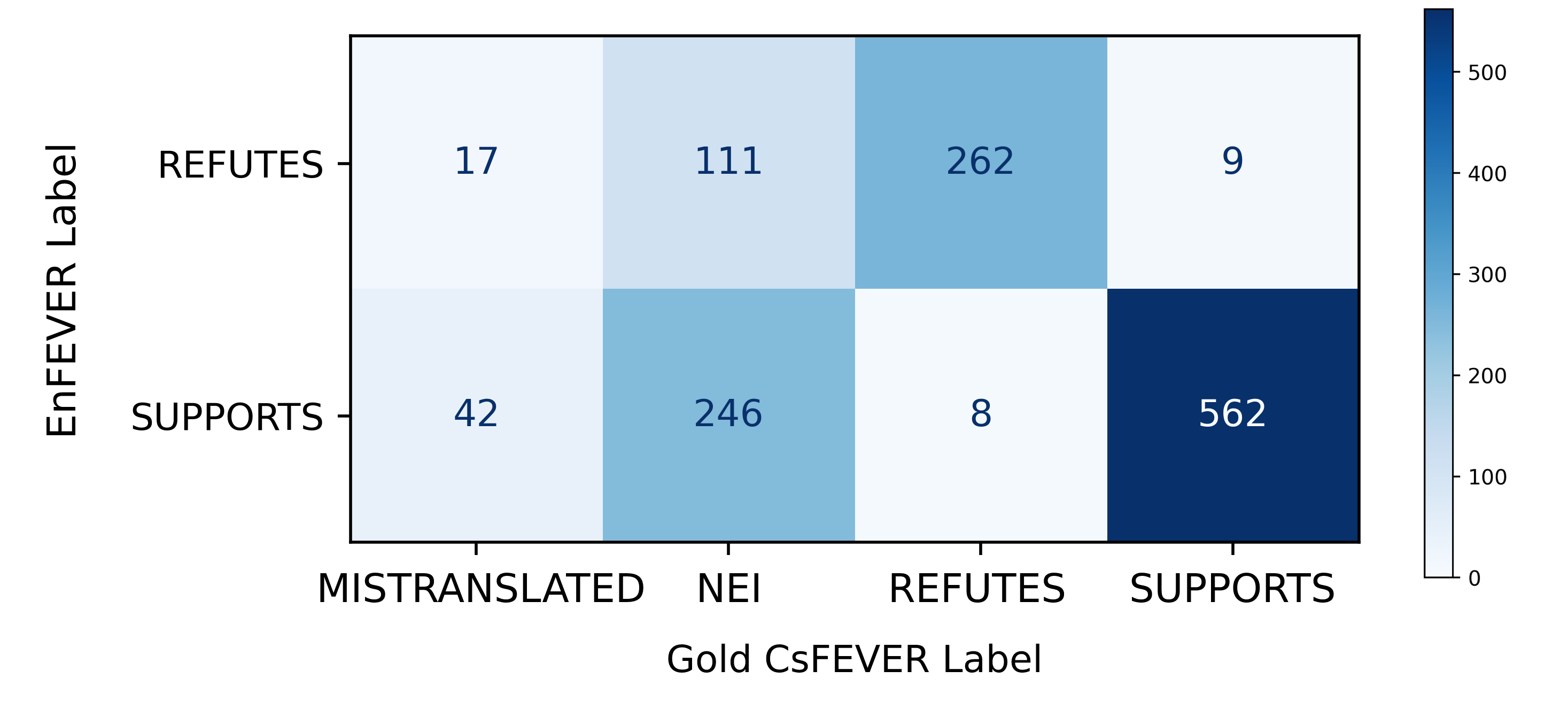}}
    \caption[Confusion matrix of the \FCZ translation scheme]{Confusion matrix of the \FCZ localization scheme }
    \label{fig:csfever-confmat}
\end{figure}

Overall, we have measured a 66\% transduction precision with a confusion distribution visualised in Figure~\ref{fig:csfever-confmat} -- 28\% of our \FCZ sample pairs were invalid due to \texttt{NOT ENOUGH INFO} in the proposed Czech \Wikipedia abstracts, 5\% sample claims were invalidated by an inadequate translation.

%--- FIG: UTF forms
 %confmatky
    %--- /FIG

We, therefore, claim that the localization method, while yielding mostly valid datapoints, needs a further refinement, and the \FCZ as-is is noisy and mostly appropriate for experimental benchmarking of model recall in the document-level retrieval task.
\revision{In Section~\ref{sec:dict} we proceed to use this data for training Czech retrieval models for the task of \textit{dictionary generation} vital for our \CTK annotations.}
With caution, it may also be used for NLI experiments. 

We conclude that while the large scale of the obtained data may find its use, a collection of novel Czech-native dataset is desirable for finer tasks, and we proceed to annotate a \CTK dataset for our specific application case. 
\revision{As Figure~\ref{fig:csfever-confmat} shows a common problem with \NEI{} mislabeling, the dataset could also be further cleaned by a well-performing NLI model at an appropriate level of confidence.}

\subsection{\revision{\FCZNLI}}\label{sec:fcznli}

\revision{Alternative way to look at the \FEN data is to view them as \textit{context}-\textit{query} pairs, where \textit{query} is a claim, and \textit{context} is a concatenation of the full texts of its evidence.
This was examined for the NLI task in~\cite{nie2019combining}, and released as the \FEVERNLI dataset.
Where no context was given (\NEI{} datapoints without evidence), the authors uniformly sampled 3--5 sentences from the top-ranked  \Wikipedia abstract according to their retrieval model.

This interpretation of the \FEN data reduces the size of \Wikipedia{} content that needs to be translated alongside the claims to 15M words  at the cost of the ability to use the data for retrieval tasks.
Therefore, we also publish a dataset we call \FCZNLI that was generated independently on the scheme from Section~\ref{fcz-method} by directly translating 228k \FEVERNLI pairs published in~\cite{nie2019combining} using \textsc{DeepL}.
We conclude that by only using the relevant parts of English \Wikipedia and translating these, we mitigate the problems found in Section~\ref{fcz-validity} and provide a solid dataset for the NLI task on the fact-checking pipeline.
}

\begin{comment}
    TODO: 
    The first dataset, CsFEVER is probably not a trustworthy localization of
the FEVER dataset due to the chosen localization methodology. By not directly
translating the ground truths and just switching to the Czech Wiki
counterparts, sentence-to-sentence alignment, a crucial requirement for the
fact-checking task, is lost. Moreover, Section 3.2.1 shows that
a lot of information is also lost, ending up and NOT ENOUGH INFO.

The issue is addmitted: "We, therefore, claim that the localization method,
while yielding mostly valid datapoints, needs a further refinement" and yet no
refinement is proposed. A "workaround" is offered which is in fact not
a workaround because it is done on a subset of data for a different task (NLI).

Finally, the entire Czech localization is dismissed as: "We conclude that while
the large scale of the obtained data may find its use, a collection of novel
Czech-native dataset is desirable for finer tasks, and we proceed to annotate
a CTKFacts dataset for our specific application case." I do not know what to
make of such conclusion. Does it make CsFEVER useless? For which tasks it is
useful then? Can it be safely considered a Czech localization of FEVER or not?

\end{comment}
%!TEX ROOT=../lrec2021.tex
\section{CTKFacts}\label{sec:ctkfacts}
In this section, we address collection and analysis of the \CTK dataset - our novel dataset for fact verification in Czech.
The overall approach to the annotation is based on \FEVER~\cite{fever2018}.
Unlike other \FEVER-inspired datasets~\cite{aly2021feverous,norregaard2021danfever,schuster-etal-2021-vitaminc} which deal with corpora of encyclopedic language style, \CTK uses a ground truth corpus extracted from an archive of press agency articles. 
% -- a considerably different language form. 

As the CTK archive is proprietary and kept as a trade secret, the full domain of all possible evidence may not be disclosed.
Nevertheless, we provide public access to the derived NLI version of the \CTK dataset  we call \CTKNLI.
\CTKNLI is described in Section~\ref{sec:ctknli}.

\subsection{CTK corpus}
For the ground truth corpus, we have obtained a proprietary archive of the Czech News Agency,\footnote{\url{https://www.ctk.eu/}} also referred to as CTK, which is a public service press agency providing news reports and data in Czech to subscribed news organizations.
Due to the character of the service -- that is, providing raw reports that are yet to be interpreted by the commercial media -- we \revision{hypothesize} such corpus suffers from significantly less noise in form of sensational headlines, political bias, etc.

\revision{Using news corpus as a ground-truth database might be (rightfully) considered controversial.
We stress that it is important to select only highly reliable sources for this purpose.
Specifically, in the Czech media environment, the CTK is known to keep the high standard of news verification.\footnote{See Czech News Agency codex at \url{https://www.ctk.cz/o_ctk/kodex/} (Czech only).}} 

The full extent of data provided to our research is 3.3M news reports published between 1 January 2000 and 6 March 2019.
We reduce this number by neglecting redundancies and articles formed around tables (e.g., sport results or stock prices).
\begin{comment} TOFINALIZE:
     "This, however, has yet to be checked." I believe it should have
been checked already, but let us be honest, it is a matter of perspective and
difficult to check. Why not change "we argue that" to less strict "we
hypothesize that" and drop the "has yet to be checked" (let us not rush into
difficult problems).
\end{comment}
Ultimately, we arrive to a corpus of 2M articles with a total of 11M paragraphs.
Hereinafter, we refer to it as to the \textit{CTK corpus}, and it is to be used as the verified text database for our annotation experiments.

\subsection{Paragraph-level documents}
The \FEVER shared task proposed a \textit{two-level} retrieval model: first, a set of \textit{documents} (i.e., \Wikipedia abstracts) is retrieved. These are then fed to the \textit{sentence retrieval} system which provides the evidence on the sentence level.

This two-stage approach, however, does not match properties of the news corpora -- in most cases, the news sentences are significantly less \textit{self-contained} than those of encyclopedic abstract, which disqualifies the sentence-level granularity.

On the other hand, the news articles tend to be too long for many of the state-of-the-art \textit{document retrieval} methods.
\FEVER addresses a similar issue by trimming the articles to their short abstracts only.
\revision{Such a trimming can not be easily applied to our data, as the news reports come often without abstracts or summaries and scatter the information across all their length.}

In order to achieve a reasonable document length, as well as to make use of all the information available in our corpus, we opt to work with our full data on the \textit{paragraph} level of granularity, using a single-stage retrieval.
From this point onwards, we refer to the CTK paragraphs also as to the \textit{documents}.

We store meta-data for each paragraph, identifying the article it comes from, its order\footnote{Headline    being considered as the 0-th paragraph.} and a timestamp of publication.

\subsection{Source Document Preselection}
\label{sec:preselection}
In \FEVER, every claim is \revision{derived from} a random sentence of a \Wikipedia article abstract sampled from the fifty thousand most popular articles~\cite{fever2018}.

With the news report archive in its place, the approach does not work well, as most paragraphs do not contain any check-worthy information.
In our case, we were forced to include an extra manual preselection task (denoted \Tzero, see Section~\ref{sec:annotation}) to deal with this problem.

\subsection{\textit{Dictionary} Generation}
\label{sec:dict}

In \FEN \textit{Claim Extraction} as well as in the annotation of \FDAN~\cite{binau2020danish}, the annotator is provided with a source \Wikipedia abstract and a \textit{dictionary} composed of the abstracts of pages \textit{hyperlinked} from the source.
The aim of such dictionary is to 1) introduce more information on entities covered by the source, 2) extend the context in which the new claim is extracted in order to establish more complex relations to other entities.

With the exception of the \textit{claim mutation} task (see below), annotators are instructed to disregard their own world knowledge.
The \textit{dictionary} is essential to ensure that the annotators limit themselves to the facts \mbox{(dis-)provable} using the corpus while still having access to higher-level, more interesting entity relations.

As the CTK corpus (and news corpora in general) does not follow any rules for internal linking, it becomes a significant challenge to gather reasonable dictionaries.
The aim is to select a relatively limited set of documents to avoid overwhelming the annotators.
These documents should be highly relevant to the given \textit{knowledge query}\footnote{Claim or paragraph formed in natural language accompanied by a timestamp of a date at which the knowledge should hold.} while covering as diverse topics as possible at the same time to allow complex relations between entities.

Our approach to generating dictionaries combines NER-augmented keyword-based document retrieval method and a semantic search followed by clustering to promote diversity.

The keyword-based search uses the TF-IDF \DrQA~\cite{chen2017drqa} document retrieval method being a designated baseline for the \FEN~\cite{fever2018}.
Our approach makes multiple calls to the \DrQA, successively representing the query $q$ by all possible pairs of \textit{named entities} extracted from the $q$.
As an example consider the query $q=\text{\enquote{Both Obama and Biden visited Germany.}}$: 
$N=\{\text{\enquote{Obama}, \enquote{Biden}, \enquote{Germany}}\}$ is the extracted set of top-level named entities.
\DrQA is then called $\binom{\abs{N}}{2} = 3$ times for the keyword queries $q_1 = \text{\enquote{Obama, Biden}}$, $q_2 = \text{\enquote{Obama, Germany}}$, $q_3 = \text{\enquote{Biden, Germany}}$.

Czech Named Entity Recognition is handled by the model of~\cite{strakova2019}.
In the end, we select at most $n_\text{KW}$ (we use $n_\text{KW} = 4$) documents having the highest score for the dictionary.

This iterative approach aims to select documents describing mutual relations between pairs of NERs.
It is also a way to promote diversity between the dictionary documents. 
Our initial experiments with a naïve method of simply retrieving documents based on the original query $q$ (or simple queries constructed from all NERs in $N$) were unsuccessful as journalists often rephrase, and the background knowledge can be found in multiple articles.
Hence, the naïve approach often reduces to search for these rephrased but redundant textual segments.
% \jd{Discuss why not include higher-level relations, e.g., NER triples.}

The second part of the \textit{dictionary} is constructed by means of semantic document retrieval.
We use the \MBERT~\cite{devlin2019bert} model finetuned on \FCZ (see Section~\ref{section:dr-baseline}), which initially retrieves rather large set of $n_\text{PRE} = 1024$ top ranking documents for the query $q$.
In the next step, we cluster the $n_\text{PRE}$ documents based on their \texttt{[CLS]} embeddings 
%\begin{comment}(\todo{\jd{CHECK + [REF!]}})\end{comment}
using $k$-means.
Each of the $k$ ($k=2$ in our case) clusters then represents a semantically diverse set of documents (paragraphs) $P_i$ for $i \in \{1, \ldots, k\}$.
Finally, we cyclically iterate through the clusters, always extracting a single document $p \in P_i$ closest to $q$ by means of the cosine similarity, until the target number of $n_\text{SEM}$ (we used $n_\text{SEM} = 4$) documents is reached.
The final dictionary is then a union of $n_\text{KW}$ and  $n_\text{SEM}$ documents selected by both described methods.

During all steps of dictionary construction, we make sure that all the retrieved documents have an older timestamp than the source.
Simply put, to each query, we assign a date of its formulation, and only verify it using the news reports published \textit{to that date}. 
The combination of the keyword and semantic search, as well as the meta-parameters involved, are a result of empirical experiments.
They are intended to provide a minimum neccessary context on the key actors of the claim and its nearest semantical neighbourhood.

In the following text, we denote a dictionary computed for a query $q$ as $d(q)$.
In the annotation tasks, it is often desirable to combine dictionaries of two different queries (claim and its source document) or to include the source paragraph itself.
For clarity, we use the term \textit{knowledge scope} to refer to such entire body of information.

\subsection{Annotation Workflow}
\label{sec:annotation}
The overall workflow is depicted in~\ref{fig:antasks} and described in the following list:

\begin{figure}
    \centering
    \includegraphics[width=\textwidth]{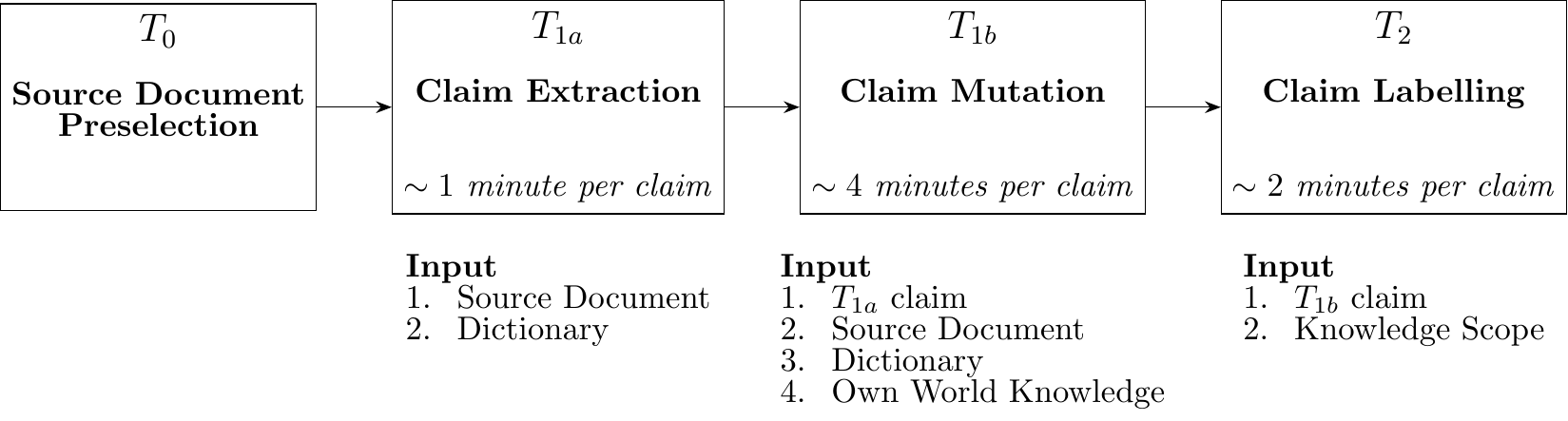}
    \caption{CTKFacts annotation tasks.}\label{fig:antasks}
\end{figure}

\begin{enumerate} 
\item\textbf{Source Document Preselection (\Tzero)} is the preliminary annotation step as described in Section~\ref{sec:preselection}, managed by the authors of this paper.

\item\textbf{Claim Extraction (\ToneA):}
\begin{itemize}
    \item The system samples random paragraph $p$ from the set of paragraphs preselected in the \Tzero~stage.
    \item The system generates a dictionary $d(p)$, querying for the paragraph $p$ and its publication timestamp (see Section~\ref{sec:dict}).
    \item Annotator $A$ is provided the \textit{knowledge scope} $K = \{p\}\cup d(p)$.
    \item $A$ is further allowed to augment $K$ by other paragraphs published in the same article as some paragraph already in $K$, in case the provided knowledge needs reinforcement.
    \item $A$ outputs a simple true \textit{initial claim} $c$ supported by $K$ while disregarding their own world knowledge.
\end{itemize}
\item\textbf{Claim Mutation (\ToneB):} The claim $c$ is fed back to $A$, who outputs a set of claim mutations: $M=\{m_1,\dots m_n\}$.
These involve the mutation types defined in~\cite{fever2018}: \textit{rephrase}, \textit{negate}, \textit{substitute similar}, \textit{substitute dissimilar}, \textit{generalize}, and \textit{specify}.
We use the term \textit{final claim} interchangeably with the \textit{claim mutation} in the following paragraphs.
This is the only stage where $A$ can employ own world knowledge, although annotators are advised to preferably introduce knowledge that is likely to be covered in the corpus. 
To catch up with the additional knowledge introduced by $A$, the system precomputes dictionaries $d(m_1),\dots d(m_n)$.

\item\textbf{Claim Labeling (\Ttwo):} 
\begin{itemize}
    \item The annotation environment randomly samples a \textit{final claim} $m$ and presents it to  $A$ with a knowledge scope $K$ containing the original source paragraph $p$, its \ToneA{} dictionary $d(p)$, as well as the additional dictionary $d(m)$ retrieved for $m$ in \ToneB{}. The order of $K$ is randomized (except for $p$ which is always first) not to bias the time-constrained $A$.
    \item $A$ is further allowed to augment $K$ by other paragraphs published in the same article as some paragraph already in $K$, in case the provided knowledge needs reinforcement
    \item $A$ is asked to spend $\leq 3$ minutes looking for minimum evidence sets  $E^{m}_1,\dots,E^{m}_n$ sufficient to infer the veracity label which is expected to be the same for each set.
    %\hu{We do not use $g()$ anywhere else, but i thought it could be helpful to understand that the evidence sets are considered independent and self-contained, just having the same veracity label}
    \item If none found, $A$ may also label the $m$ as \NEI.
\end{itemize}
\end{enumerate}

Note that \FEVER defines two subtasks only: \textit{Claim Generation} and \textit{Claim Labeling}. 
\textit{The Claim Generation} corresponds to our \Tone, while \textit{the Claim Labeling} is covered by \Ttwo.

\subsubsection{Annotation Platform}
\label{fcheck}
Due to notable differences in experiment design, we have built our own annotation platform, rather than reusing that of~\cite{fever2018}. 
The annotations were collected using a custom-built web interface.
Our implementation of the interface and backend for the annotation workflow described in Section~\ref{sec:annotation} is distributed under the MIT license and may be inspected online.\footnote{\url{https://fcheck.fel.cvut.cz}}
We provide further information on our annotation platform in appendix~\ref{appendix:annotations}.

\subsubsection{Annotators}
Apart from \Tzero, the annotation tasks were assigned to groups of bachelor and master students of Journalism from the Faculty of Social Sciences at the Charles University in Prague.
We have engaged a total of 163 participants who have signed themselves for courses in \textit{AI Journalism} and \textit{AI Ethics} during the academic year 2020/2021.
We used the resulting data, trained models and the annotation experiment itself to introduce various NLP mechanisms, as well as to obtain valuable feedback on the task feasibility and pitfalls.

The annotations were made in several \textit{waves} -- instances of the annotation experiment performed with different groups of students.
This design allowed us to adjust the tasks, fullfilment quotas and the interface after each wave, iteratively removing the design flaws.

\subsubsection{Cross-annotations}
\label{sec:cross-annotations}
In the annotation labeling task, we advised the annotators to spend no more than 2-3 minutes finding as many evidence sets as possible within \Wikipedia, so that the dataset can later be considered almost \textit{exhaustive}~\cite{fever2018}.
With our CTK corpus, the exhaustivity property is unrealistic, as the news corpora commonly contain many copies of single ground truth. 
For example, claim \enquote{\textit{Miloš Zeman is the Czech president}} can be supported using any \enquote{\textit{\enquote{\dots}, said the Czech president Miloš Zeman.}} clause occurring in corpus.

Therefore, we propose a different scheme: annotator is advised to spend 2-3 minutes finding as many distinct evidence sets as possible within the time needed for good reading comprehension.
%Median yield, however, was 1 evidence set per annotation.
Furthermore, we have collected an average of 2 cross-annotations for each claim.
This allowed us to merge the evidence sets across different \Ttwo{} annotations of the same claim, as well as it resulted in a high coverage of our cross-validation experiments in Section~\ref{sec:agreement}.

\subsection{Dataset analysis and postprocessing}

\begin{table}
    \begin{center}
        \begin{minipage}{\textwidth}
            \caption{Label distribution in \CTK splits before and after cleaning.}\label{tab:ctkfacts-overview}
            \begin{tabular*}{\textwidth}{@{\extracolsep{\fill}}lllllll@{\extracolsep{\fill}}}
                \toprule
                & \multicolumn{3}{@{}c@{}}{\CTK uncleaned, balanced} & \multicolumn{3}{@{}c@{}}{\CTK cleaned, stratified}\\
                \cmidrule{2-4}\cmidrule{5-7}
                Split & {\texttt{SUPPORTS}} & \texttt{REFUTES}  & \texttt{NEI} & {\texttt{SUPPORTS}} & \texttt{REFUTES}  & \texttt{NEI}\\
                \midrule
                \train  & 1,164 & 549 & 503     & 1,104 & 556 & 723 \\
                \dev    & 100 & 100 & 100       & 142 & 85 & 105\\
                \test   & 200 & 200 & 200       & 176 & 79 & 127\\
                \botrule
            \end{tabular*}
        \end{minipage}
    \end{center}
\end{table}

\label{sec:dataset}
After completing the annotation runs, we have extracted a total of 3,116 multi-annotated claims.
47\% were \texttt{SUPPORT}ed by the majority of their annotations, \REF{} and \NEI{} labels were approximately even, the full distribution of labels is listed in Table~\ref{tab:ctkfacts-overview}.
% We have originally experimented with balanced \dev and \test splits to punish predictors exploiting this bias.

Of all the annotated claims, 1,776, that is 57\%, had at least two independent labels assigned by different annotators.
This sample was given by the intrinsic randomness of \Ttwo{} claim sampling.
In this section, we use it to asses the quality of our data and ambiguity of the task, as well as to propose annotation cleaning methods used to arrive to our final \text{cleaned} \CTK dataset.

\subsubsection{Inter-Annotator Agreement}
\label{sec:agreement}
\begin{comment}
    TOFINALIZE: Section 4.6.1 (4.6.1 Inter-Annotator Agreement): Maybe you would like to
include the inter-annotator agreement values from other similar publications
for comparison? Also, why Krippendorff's alpha rather than Fleiss kappa used by
FEVER (Thorne et al., 2018)? If missing observations do not allow for Fleiss
kappa, can you measure it at least on a subset of annotations on which the
annotators intersect?

\end{comment}
Due to our cross-annotation design (Section~\ref{sec:cross-annotations}), we had generously sized sample of independently annotated labels in our hands.
As the total number of annotators was greater than 2, and as we allowed missing observations, we have used the Krippendorff's alpha measure~\cite{krippendorff1970} which is the standard for this case~\cite{hayes2007krippendorff}.
\revision{For the comparison with \cite{fever2018} and \cite{norregaard2021danfever}, we also list a 4-way Fleiss' $\kappa$-agreement~\cite{fleiss1971measuring} calculated on a sample of 7.5\% claims.}
%todo mention theirs

We have calculated the resulting Krippendorff's alpha agreement to be 56.42\% \revision{and Fleiss' $\kappa$ to be 63\%.}
We interpret this as an adequate result that testifies to the complexity of the task of news-based fact verification within a fixed knowledge scope.
It also encourages a round of annotation cleaning experiments that would exploit the number of cross-annotated claims to remove common types of noise.

\subsubsection{Manual annotation cleaning}
\label{sec:cleaning}

We have dedicated a significant amount of time to manually traverse \textit{every} conflicting pair of annotations to see if one or both violate the annotation guidelines.
The idea was that this should be a common case for such annotations, as the CTK corpus does not commonly contain a conflicting pair of paragraphs except for the case of \textit{temporal reasoning} explained in Section~\ref{sec:annot_problems}.

After separating out 14\% (835) erroneously formed annotations, we have been able to resolve every conflict, ultimately achieving a full agreement between the annotations.
We discuss the main noise patterns in Section~\ref{sec:annot_problems}.

\subsubsection{Model-based annotation cleaning}
\label{sec:misclas-cleaning}
Upon evaluating our NLI models (Section~\ref{sec:nli-baseline}), we have observed that model misclassifications frequently occur at \Ttwo{} annotations that are counterintuitive for human, but easier to predict for a neural model.

Therefore, we have performed a series of experiments in model-assisted human-in-the-loop data cleaning similar to~\cite{guyon1994patterns} in order to catch and manually purge outliers, involving an expertly trained annotator working without a time constraint:

\begin{enumerate}
    \item A \textit{fold} of dataset is produced using the current up-to-date annotation database, sampling a stratified \test split from all untraversed claims - the rest of data is then divided into \dev and \train stratified splits, so that the overall train-dev-test ratio is roughly 8:1:1.
    \item Mark the \test claims as traversed.
    \item A round of NLI models (Section~\ref{sec:nli-baseline}) is trained \revision{using the current \train split} to obtain the strongest veracity classifier for the current fold. 
    The individual models are optimized w.r.t. the \dev split, while the strongest one is finally selected using \test. 
    \item \test-misclassifications of this model are then presented to an expert annotator along with the model suggestion and an option to remove an annotation violating the rules and to propose a new one in its place.
    \item New annotations propagate into the working database and while there are untraversed claims, we proceed to step 1.
\end{enumerate}

Despite allowing several inconsistencies with the scheme above during the first two folds (that were largely experimental), this led to a discovery of another 846 annotations conflicting the expert annotator's labeling and a proposal of 463 corrective annotations (step 4.).

%!TEX ROOT=../lrec2021.tex
\subsection{Common annotation problems}

\label{sec:annot_problems}
In this section, we give an overview of common misannotation archetypes as encountered in the cleaning stage (sections~\ref{sec:cleaning} and~\ref{sec:misclas-cleaning}).
These should be considered when designing annotation guidelines for similar tasks in the future.
The following list is sorted by decreasing appearance in our data.
\begin{enumerate}
    \item \textbf{Exclusion misassumption} is by far the most prevalent type of misannotation.
    The annotator wrongly assumes that an event connected to one entity implies that it cannot be connected to the other entity.
    E.g., evidence \textit{\enquote{Prague opened a new cinema.}} leads to \textit{\enquote{Prague opened a new museum.}} claim to be refuted.
    In reality, there is neither textual entailment between the claims, nor their negations.
    We attribute this error to confusing the \Ttwo~with a \textit{reading comprehension}\footnote{The corresponding reading comprehension task might be: \textit{\enquote{Does the article tell us that Prague opened a museum? Highlight the relevant information.}}} task common for the field of humanities.
    % We have reduced the frequency of this misclassification by introducing a \textit{golden rule} (Section~\ref{sec:golden-rules}) for it, keeping it on annotator's mind at all times

    % INFO: We need some statistics to show this
    % \textbf{\NEI{} avoidance}  -- \textit{\enquote{Pandas are endangered.}} was used once to \texttt{SUPPORT} and once to \texttt{REFUTE} the claim \textit{\enquote{Koalas are endangered.}}, zero times as \texttt{NEI}.
    % This, among other examples, shows that our annotators often preferred the \textit{definite} labels, even where \texttt{NEI} is appropriate, which might justify its underrepresentation shown in Section~\ref{sec:dataset}.

    \item \textbf{General misannotation}: we were unable to find exact explanations for large part of the mislabelled claims.
    We traced the cause of this noise to both unclearly formulated claims and UI-based user errors.
    \item \textbf{Reasoning errors} cover failures in assessment of the claim logic, e.g., confusing \enquote{less} for \enquote{more}, etc.
    Also, this often involved errors in temporal reasoning, where an annotator submits a dated evidence that contradicts the latest news w.r.t. the timestamp. 

    \item \textbf{Extending minimal evidence}: larger than minimal set of evidence paragraphs was selected. This type of error typically does not lead to misannotation, nevertheless, it was common in the sample of dataset we were analyzing. 
    \item \textbf{Insufficient evidence} where the given evidence misses vital details on entities. As an example: the evidence \textit{\enquote{A new opera house has been opened in Copenhagen.}} does not automatically support the claim \textit{\enquote{Denmark has a new opera house.}} if another piece of evidence connecting \textit{Copenhagen} and \textit{Denmark} is not available.
    This type of error indicates that the annotator of the claim extended the allowed \textit{knowledge scope} with his/hers own world-knowledge.
    
    % \item {\textbf{Mutation vagueness}} -- \textbf{Claim} fault. Mutation generalizes out an integral part of the original claim, typically the named entities. E.g., $m=\text{\enquote{The convoy is 200 metres long.}}$ \todo{Give the original claim.}

\end{enumerate}
\begin{comment}
\item \todo{Remove (A*) ids before submit.}
\item \todo{Above, we should mention that the knowledge scopes include the timestamps and define the $timestamp(m)$ function!}
\end{comment}

\subsection{\CTKNLI Dataset}
\label{sec:ctknli}
Finally, we publish the resulting cleaned \CTK dataset consisting of 3,097 manually labeled claims, with label distribution as displayed in Table~\ref
{tab:ctkfacts-overview}.
We opt for stratified splits due to the relatively small size of our data and make sure that no CTK source paragraph was used to generate claims in two different splits, so as to avoid any \textit{data leakage}.

The full CTK corpus cannot be, unfortunately, released publicly.
Nevertheless, we extract all of our 3,911 labeled \textit{claim}-\textit{evidence} pairs to form the \CTKNLI dataset.
Claim-wise, it follows the same splitting as our DR dataset, and the \NEI{} evidence is augmented by the paragraph that was used to derive the claim to enable inference experiments. 

We have acquired the authorization from CTK to publish all evidence plaintexts, which we include in \CTKNLI and open for public usage.
The dataset is released publicly on HuggingFace dataset hub\footnote{\url{https://huggingface.co/datasets/ctu-aic/ctkfacts_nli}} and provides its standard usage API to encourage further experiments.
\section{Spurious Cue Analysis}\label{sec:analysis}
% \begin{itemize}
    % \item \todo{Nevyhodime wordpiece? Otazkou je, jak dobre koreluje s predponami jako \q{ne-}. Co pouzit POS tagging? To je ale asi na dalsi paper...}
    % \item \jd{Ten \q{Maintaining} paper obsahuje hrozne chyb a nepresnosti. Mam podezreni, ze neprosel poradne recenznim rizenim. Trochu lepsi je ta diplomka \q{Danish Fact Verification}, tam ale neni DCI. \textbf{DCI zatim vypoustim. Prinasi neco duleziteho?}. Rovnez vyhazuju \textit{utility} -- skaluje pouze dolni hranici intervalu (\todo{probrat s Martinem!}).}
    % \item \jd{Maintaining paper: \q{surface-level linguistic patterns that \textit{leak} class information}}
    % \item \jd{Maintaining paper: \q{A note regarding language: in this case, we con- sider 1, 2, and 3-grams, with skips in the range of [0, 2]. This is suitable for English; other languages might benefit from broader skip ranges.}}
% \end{itemize}

\label{section:claim-quality}
In the claim generation phase \ToneB, annotators are asked to create mutations of the initial claim. 
These mutations may have a different truth label than the initial claim or even be non-verifiable with the given knowledge database.
During trials in~\cite{fever2018}, the authors found that a majority of annotators had a difficulty with creating non-trivial negation mutations beyond adding \enquote{not} to the original.
Similar \textit{spurious cues} may lead to models cheating instead of performing proper semantic analysis.
    
In~\cite{binau2020danish}, the authors investigated the impact of the trivial negations on the quality of the \FEN and \FDAN datasets.
Here we present similar analysis based on the \textit{cue productivity} and \textit{coverage} measures derived from work of~\cite{niven2019probing}.

In our case the cues extracted from the claims have a form of unigrams and bigrams. 
The definition of the \textit{productivity} assumes a balanced dataset with respect to labels. 
The \textit{productivity} of a cue $k$ is calculated as follows:
\begin{equation} \label{equ:productivity-cue}
    \pi_k = \frac{\max\limits_{c \in C} \abs{A_{cue = k} \cap A_{class = c}}}{\abs{A_{cue = k}}},
\end{equation}

where $C$ denotes the set of possible labels $C=\{\SUP{}, \REF{}, \NEI{}\}$, $A$ is the set of all claims, $A_{cue = k}$ is the set of claims containing cue $k$ and $A_{class = c}$ is the set of claims annotated with label $c$. 
Based on this definition the range of \textit{productivity} is limited to $\pi_k \in [1/\abs{C}, 1]$, for balanced dataset.
% \noindent There is also a proposed metric suitable for comparison between datasets, called utility that normalizes productivity by a number of distinct labels, which may differ across datasets
% \begin{equation} \label{equ:utility-cue}
%     % \mathbb{1}
%     \rho_{k} = \pi_{k} - \frac{1}{|Y|}
% \end{equation}
The \textit{coverage} of a cue is defined as a ratio $\xi_{k} = \abs{A_{cue = k}}/\abs{A}$.

We take the same approach as~\cite{binau2020danish} to deal with the dataset imbalance: the resulting metrics are obtained by averaging over ten versions of the data based on random subsampling. 
We compute the metrics for both \FCZ and \CTK datasets.
% In addition to unigram and bigram cues for the productivity and coverage metrics, we also tried to use lower-granular worpiece tokens as cues.
% Regarding the DCI, we used wordpieces, unigrams and 4-skip-bigrams as cues.
Similarly to~\cite{binau2020danish}, we also provide the harmonic mean of productivity and coverage, which reflects the overall effect of the cue on the dataset.

The results in Table~\ref{table:claim-quality-fever-prod-cov} show that the cue bias detected in \FEN claims~\cite{binau2020danish} propagates to the translated \FCZ, where the words \q{není} (\q{is not}) and \q{pouze} (\q{only}) showed high productivity of 0.57 and 0.55 and ended in the first 20 cues sorted by the \emph{harmonic mean}.
However, their impact on the quality of the entire dataset is limited as their coverage is not high, which is illustrated by their absence in the top-5 most influential cues. 
Similar results for the \CTK are presented in Table~\ref{table:claim-quality-ctk-prod-cov}.

% According to the \emph{harmonic mean}, when using wordpiece tokens, the most influential are \enquote{\#\#'}, which is accent at the end of the word token, and \enquote{UNK}, which is a special token that includes any token not found in the dictionary (see Table~\ref{table:claim-quality-fever-dci}).
% Despite the fact that they provide very little information to the model, they hold a dominant position in the results due to their high occurrence.

% The results on the ČTK dataset are significantly affected by the fact that the number of claims is quite low.
% This causes that even specific cues based on the thematic cluster formed around the original statement have a relatively higher impact on the dataset (for example, \q{Bühler Motor} in the Table~\ref{table:claim-quality-ctk-dci}).
% Although ČTK also contains some constructs with a higher productivity, for example \q{Thomas Alva} (0.69) or "není (0.7), their influence on the whole dataset is negligible.
% While the analysis did not confirm any significant bias in the claims, there is still a need to monitor these metrics in the future as more claims are made.

\begin{table}
\begin{center}
\begin{minipage}{0.93\textwidth}
\caption{Productivity, coverage and their harmonic mean calculated on \FCZ dataset claims sorted by the decreasing harmonic mean.}
\label{table:claim-quality-fever-prod-cov}
\begin{tabular*}{\textwidth}{lllrrrr}
\toprule
rank & cue\textsubscript{cs} & cue\textsubscript{translated} & label & productivity &  coverage &  h. mean \\
\midrule
\multicolumn{6}{c}{Unigrams} \\
1 & je &        is &           SUP &          0.34 &      0.24 &           0.28 \\
2 & v &      in/at &           REF &          0.35 &      0.20 &           0.25 \\
3 & se &           &           REF &          0.36 &      0.15 &           0.21 \\
4 & byl &      was &           NEI &          0.37 &      0.09 &           0.14 \\
5 & na &        on &           NEI &          0.36 &      0.08 &           0.13 \\
\midrule
\multicolumn{6}{c}{Unigram negations} \\
21 & není &      is not &             REF &          0.91 &      0.02 &           0.04 \\
61 & nebyl &     wasn't &              REF &          0.91 &      0.01 &           0.02 \\
114 & nemá &      doesn't have &       REF &          0.87 &      0.00 &           0.01 \\
171 & nebyla &    wasn't &             REF &          0.91 &      0.00 &           0.01 \\
294 & nehrál &    didn't play &        REF &          0.91 &      0.00 &           0.00 \\
\midrule
\multicolumn{6}{c}{Bigrams} \\
1 & v roce &       in year &             REF &          0.45 &      0.06 &           0.11 \\
2 & ve filmu &     in movie &            SUP &          0.46 &      0.04 &           0.07 \\
3 & se narodil &   was born &            REF &          0.46 &      0.02 &           0.04 \\
4 & Ve filmu &     In movie &            NEI &          0.48 &      0.02 &           0.03 \\
5 & se nachází &   is located &          REF &          0.41 &      0.01 &           0.03 \\
\botrule
\end{tabular*}
\end{minipage}
\end{center}
\end{table}

\begin{table}
\begin{center}
\begin{minipage}{0.93\textwidth}
\caption{Productivity, coverage and their harmonic mean calculated on ČTK dataset claims sorted by the harmonic mean. }
\label{table:claim-quality-ctk-prod-cov}
\begin{tabular*}{\textwidth}{lllrrrr}
\toprule
rank & cue\textsubscript{cs} & translation & label & productivity &  coverage &  h. mean \\
\midrule
\multicolumn{5}{c}{Unigrams} \\
1 & v  &    in/at  &     NEI &          0.34 &      0.29 &           0.31 \\
2 & se &          &     SUP &          0.35 &      0.15 &           0.21 \\
3 & na &    on     &     SUP &          0.35 &      0.13 &           0.19 \\
4 & je &    is     &     REF &          0.37 &      0.11 &           0.17 \\
5 & V  &    In/At  &     NEI &          0.44 &      0.09 &           0.15 \\
\midrule
\multicolumn{5}{c}{Unigram negations} \\
56  & není        &     is not &           REF &        0.79 &      0.01 &           0.02 \\
96  & nesouhlasí  &     disagrees &        NEI &        0.50 &      0.01 &           0.01 \\
174 &     nebude  &     won't &            REF &        0.70 &      0.01 &           0.01 \\
218 &       nemá  &     doesn't have &     REF &        0.78 &      0.00 &           0.01 \\
696  & nelegální  &     illegal &          NEI &        0.60 &      0.00 &           0.01 \\
\midrule
\multicolumn{5}{c}{Bigrams} \\
1 & v roce &     in year &         REF &        0.35 &      0.04 &           0.07 \\
2 & se v &       in/at &           NEI &        0.40 &      0.02 &           0.03 \\
3 & V roce &     In year &         REF &        0.39 &      0.02 &           0.03 \\
4 & v Praze &    in Prague &       REF &        0.40 &      0.02 &           0.03 \\
5 & více než &   more than &       SUP &        0.41 &      0.01 &           0.02 \\
\botrule
\end{tabular*}
\end{minipage}
\end{center}
\end{table}

%!TEX ROOT=../lrec2021.tex
\section{Baseline models}\label{sec:baseline}

% \begin{itemize}
%     \item \jd{Show comparison to \FEN}
% \end{itemize}
\begin{comment}
    TOFINALIZE: Section 6: "In this section, we establish baseline models for both CsFEVER and
CTKFacts." I am not sure if baseline models can be established for CTKFacts
when nobody from the NLP community will ever be able to work on the data... or
rather, they can be established, but for what use?

\end{comment}
\revision{In this section, we explore the applicable models for both \FCZ and \CTK. 
We train a round of currently best-performing \textit{document retrieval} (DR) and \textit{natural language inference} (NLI) models, \revision{to examine the difficulty of the task and to establish a baseline to our datasets, mainly \FCZNLI and \CTKNLI}. We also give results for \FEN providing a point of reference to the well-established dataset.}

%!TEX ROOT=../lrec2021.tex

\subsection{Document Retrieval}
\label{section:dr-baseline}
We provide four baseline models for the document retrieval stage: \DrQA and \Anserini represent classical keyword-search approaches, while multilingual BERT (\MBERT) and \ColBERT models are based on Transformer neural architectures.

In line with \FEVER~\cite{fever2018}, we employ document retrieval part of \DrQA~\cite{chen2017drqa} model.
% \DrQA was designed for \textit{machine reading at scale} --- a combination of large-scale open-domain question answering and machine comprehension of text.
The model was originally used for answering questions based on \Wikipedia corpus, which is relatively close to the task of fact-checking.
The DR part itself is based on TF-IDF weighting of BoW vectors while optimized by using hashing.
We calculated the TF-IDF index using \DrQA implementation for all unigrams and bigrams with $2^{24}$ buckets.
% \jd{We should link the implementation (Github).}

Inspired by the criticism of choosing weak baselines presented in~\cite{yang2019hype}, we decided to validate our TF-IDF baseline against the proposed \Anserini toolkit implemented by Pyserini~\cite{lin2021pyserini}.

We computed the index and then finetuned the $k_1$ and $b$ hyper-parameters using grid search on defined grid $k_1$ $\in$ [0.6, 1.2], $b$ $\in$ [0.5, 0.9], both with step 0.1.
\revision{On a sample of 10,000 training claims, we selected the best performing parameter values: for \FCZ these were $k_1 = 0.9$ and $b = 0.9$, while for \FEN and \CTK we proceed with $k_1 = 0.6$ and $b=0.5$.}

% \jd{What is the exact model? I don't see any mention on "multilingual" in the Devlin's paper. EDIT: it is not mentioned, but it seems that it is the suggested reference: \url{https://github.com/google-research/bert}. Adding~\cite{pires201multilingual}.}
% \mr{It was trained later while using the exactly same architecture as the BERT uses only using multi-language source data. I have noticed that when using m-BERT they cite the Devlins paper typically, but I might be wrong}
% We used the cased multilingual  \MBERT from the HuggingFace library~\cite{wolf2020transformers} as the backbone model.

Another model we tested is the \MBERT~\cite{devlin2019bert}, which is a representative of Transformer architecture models.
We used the same setup as in~\cite{chang2020twotower} with an added linear layer consolidating the output into embedding of the required dimension~512.
In the fine-tuning phase, we used the claims and their evidence as relevant (positive) passages.
For multi-hop claims, based on combinations of documents, we split the combined evidence, so the queries are always constructed to relate to a single evidence document, only.
\revision{Unlike in~\cite{chang2020twotower}, we used a smaller training batch size of $128$ and learning rates $10^{-5}$ for the \textsf{ICT+BFS} tuning and $5\times 10^{-6}$ for the fine-tuning stage.}
% There exists claims, although relatively small amount, that were created by combining several passages from different articles (multi-hop claims).
% Without prejudice to the generality of the solution, we split the combined evidence, so query is always in a relation with only a single evidence passage.
% This way we slightly increase the amount of data by cloning the query for each part of its evidence.
%To illustrate this, consider the statement \emph{Prague is the capital of one of the European states.} with evidence combining two documents: \emph{Prague is the capital of the Czech Republic.} in a document A and \emph{Prague is a European country.} in a document B.
    
We used this fine-tuned model to generate 512-dimensional embeddings of the whole document collection.
In the retrieval phase, we used the FAISS library~\cite{johnson2019faiss} and constructed \emph{PCA384 Flat} index for \CTK and \emph{Flat} index for \FCZ data.\footnote{\revision{We do not provide \MBERT results for \FEN due to computational complexity and better results of \ColBERT demonstrated for \FCZ and \CTK (see below).}}
% In the case of the \emph{PCA384 Flat} index, the original output of the pre-trained model with dimension 512 is transformed into a 384-dimensional vector using PCA. This was done to lower the memory footprint.
% \emph{Flat} index uses the full 512-dimensional vector.
% When given a query, the FAISS index then retrieves top-k most relevant documents, implemented as k-means clustering.
\begin{comment}
\noindent \textbf{\FCZ \MBERT:} The model was pretrained on ICT and BFS tasks on the \Wikipedia articles for 20 epochs.
The training was done on 4 NVIDIA Tesla V100 GPUs with batch size 128, Adam optimizer with $10^{-5}$ learning rate and no weight decay.
In the fine-tuning phase was used the model from the pretraining phase that showed best performance on \FCZ development set.
Then it was fine-tuned for next 20 epochs with the same setup except for learning rate, which was reduced to $5\times 10^{-6}$ similarly to \cite{chang2020twotower}.
As a result, we saved the model with the highest performance on the development set.
    
\noindent \textbf{\CTK \MBERT:} We started with the pre-trained version of the \CTK \MBERT and added further 10 epochs of pre-training utilizing the ČTK corpus keeping the same setup as stated above.
In the fine-tuning phase, the model was tuned using real claims and their annotated evidence for 20 epochs with $5e\times 10^{-6}$ learning rate.
\end{comment}

% \subsubsection{\ColBERT}
% \label{sub:prop-colbert}
The last tested model was a recent \ColBERT, which provides the benefits of both cross-attention and two-tower paradigms~\cite{khattab2020colbert}.
We have employed the implementation as provided by the authors,\footnote{\url{https://github.com/stanford-futuredata/ColBERT}} changing the backbone model to \MBERT and adjusting for the special tokens.
\revision{The training batch size was 32, learning rate $3\times 10^{-6}$, we have used masked punctuation tokens, mixed precision and L2 similarity metric.}

The model was trained using triplets \textit{(query, positive paragraph, negative paragraph)} with the objective to correctly classify paragraphs using a cross-entropy loss function.
We constructed the training triplets so that the claim created by a human annotator was taken as a \textit{query}, a paragraph containing evidence as a \textit{positive} and a random paragraph from a randomly selected document as a \textit{negative} sample.
    
As already stated, For the \CTK, the number of claims is significantly lower than for \FCZ.
Therefore, we increased the number of CTK training triplets: instead of selecting negative paragraphs from a random document, we selected them from an evidence document with the condition that the paragraph must not be used directly in the evidence.
The number of training triplets was still low, so we also generated synthetic triplets as follows.
We generated a synthetic query by extracting a random sentence from a random paragraph.
A set of the remaining sentences of this paragraph were designated a \textit{positive paragraph}.
The \textit{negative paragraph} was, once again, selected as a random paragraph of a random document.
Then the title was used as a query instead of a random sentence, and a random paragraph from the article was used as a \textit{positive}.
\textit{Negative paragraph} was selected in the same way as above.
As a result, we generated about 950,000 triplets ($\approx{944,000}$ synthetic and $\approx{6,000}$ using human-created claims) for the \CTK.
% \hu{nitpick: Maybe I would use a single tilde (\enquote{$\sim$})}

% To generate more training triples, we repeated the process only with hard negatives, which were formed as non-evidence paragraphs from the document that contains the evidence. The amount of data was still unsatisfactory, especially for the ČTK dataset, so we generated synthetic triples. We extracted a random sentence from a paragraph, so that the remaining paragraph forms a positive passage and a randomly chosen paragraph forms a negative passage. As a result, we generated about 600 thousand triples ($\approx{500K}$ synthetic and $\approx{100K}$ original).

% We tried two setups here, 64-dimensional term representation with document trimming to a maximum of 180 tokens and richer 128-dimensional term representation with document trimming to a maximum of 220 tokens on \FCZ.
% For the ČTK dataset we counted only the larger version, as it proved to be more powerful. The query is always truncated to a maximum of 32 tokens. 
% Training was done on triples (see Section~\ref{sub:prop-colbert}) using two NVIDIA Tesla V100 GPUs with batch size 64, learning rate 3e-6, masked punctuation tokens, mixed precision and L2 similarity metric.
%The latter representation provided slightly better results.
    
We tried two setups here, 32 and 128 dimensional term representation (denoted \ColBERT{32} and \ColBERT{128}) with document trimming to a maximum of 180 tokens on \FEVER datasets.
% For the ČTK dataset we counted only the larger version, as it proved to be more powerful.
% The query is always truncated to a maximum of 32 tokens. 
% Training was done on a pair of NVIDIA Tesla V100 GPUs with batch size 64, learning rate $3\times10^{-6}$, masked punctuation tokens, mixed precision and L2 similarity metric.
The results are shown in Table~\ref{table:dr}.
Methods are compared by means of Mean Reciprocal Rank (MRR) given $k\in\{1, 5, 10, 20\}$ retrieved documents.
For \FCZ, the neural network models achieve significantly best results., with \ColBERT taking lead.
In case of \CTK, both \Anserini and \ColBERT are best performers.
Interrestingly, \MBERT fails in this task.
We found that this is mainly caused by \MBERT preference for shorter documents (including headings). \revision{As expected, the results for \FEN are comparable to those of \FCZ: the \Anserini performance improved, while \ColBERT performed slightly worse for the English corpus. Note that \FEN corpus is more than ten times larger (5.4M~pages) than the \FCZ one (452k~pages).}

\begin{table}
    \begin{center}
    % \begin{minipage}{0.7\textwidth}
    \caption{\revision{Document Retrieval baseline results in MRR (\%) for \FCZ, \CTK, and \FEN datasets.}}\label{table:dr}
    % \scalebox{0.85}{
    \begin{tabular}{llcccc}
    \toprule
    dataset & model & MRR@1  & MRR@5 & MRR@10 & MRR@20 \\
    \midrule
    \multirow{4}{5em}{\FCZ} & \DrQA & 31.23 & 38.55 & 39.14 & 39.38\\
     & \Anserini & 27.06 & 32.91 & 33.75 & 34.13 \\
     & \MBERT & 50.54 & 55.90 & 56.19 & 56.29 \\
     & \ColBERT{128} & \textbf{63.93} & \textbf{70.70} & \textbf{71.16} & \textbf{71.34} \\
     \midrule
     \multirow{4}{5em}{\CTK} & \DrQA & 9.26  & 14.25 & 15.28 & 15.55\\
      & \Anserini & 13.23 & \textbf{18.65} & \textbf{19.39} & \textbf{19.78} \\
      & \MBERT & 1.59 & 2.39 & 2.75 & 3.18 \\
      & \ColBERT{32} & \textbf{13.49} & 18.50 & 19.26 & 19.70 \\
      \midrule
     \multirow{4}{5em}{\FEN} & \DrQA & 27.09 & 36.06	 & 37.51 & 38.08\\
      & \Anserini & 32.87 & 41.48 & 42.44 & 42.91 \\
    %   & \MBERT & - & - & - & - \\
      & \ColBERT{128} & \textbf{53.57} & \textbf{64.03} & \textbf{64.73} & \textbf{64.94} \\
    \bottomrule
    \end{tabular}
    % }
    % \end{minipage}
    \end{center}
\end{table}

\subsection{Natural Language Inference}
\label{sec:nli-baseline}

\revision{The aim of the final stage of the fact-checking pipeline, the NLI task, is to classify the veracity of a claim based on the previously retrieved evidence.
We have \revision{fine-tuned} several different \revision{Czech-compatible} Transformer models \revision{on our data} in order to provide a strong baseline for this task.}

From the multilingual models, we have experimented with \SlavicBERT and \SMBERT~\cite{reimers2019sentence} models in their \textit{cased} defaults, provided by the \textsf{DeepPavlov} library~\cite{burtsev2018deeppavlov}, as well as with the original \MBERT from~\cite{devlin2019bert,pires201multilingual}.

We have further examined two pretrained \XLM-large models, one fine-tuned on an NLI-related \textit{SQuAD2}~\cite{rajpurkar2016squad} \textit{down-stream task}, other on the crosslingual \textit{XNLI}~\cite{conneau2018xnli} task.
These were provided by Deepset.\footnote{\url{https://www.deepset.ai}} and HuggingFace\footnote{\url{https://huggingface.co}} 

Finally, we have performed a round of experiments with a pair of recently published Czech monolingual models.
\RobeCzech~\cite{straka2021robeczech} was pretrained on a set of currated Czech corpora using the \RoBERTa-base architecture.
\FERNETC~\cite{DBLP:journals/corr/abs-2107-10042fernet} was pretrained on a large crawled dataset, using \BERT-base architecture. 

\revision{We have fine-tuned the listed models using 4 fact-checking datasets adapted for the NLI task using their gold sets of evidence.
\CTKNLI uses our data collected in Chapter~\ref{sec:ctkfacts} as pairs of strings containing the claim and its concatenated evidence (or the text of its \textit{source paragraph} for \NEI{} claims).
\FEVERNLI was extracted from the original \FEN dataset in~\cite{nie2019combining}, obtaining the single-string context by concatenating the evidence sentences for each claim, or a sample of 3-5 results of the proposed DR model for the \NEI{} claims.
\FCZNLI is its full Czech translation (see Section~\ref{sec:fcznli}).
\FCZ (\textit{NearestP}) was obtained from the experimental dataset extracted in Chapter~\ref{sec:fevercs} using the entire Wiki abstracts for evidence, non-verifiable claim evidence was supplemented with the top result of our \DrQA model established in Section~\ref{section:dr-baseline}.}

\revision{For each dataset, we have fine-tuned all listed models using the \texttt{sentence\_transformers} implementation of the Cross-encoder~\cite{reimers2019sentence} with two texts on input (concat-evidence, claim) and a single output value. We have experimented with multiple batch sizes for each model (varying from 2 to 10) and trained each using the Adam optimizer with $2 \cdot 10^{-5}$ default learning rate and weight decay of 0.01.
The number of linear warmup steps was determined by 10-30\% of the \train size, varying per experiment.
Ultimately, we kept the best performing models in terms of \dev accuracy for each model-dataset pair.
}

\revision{We then evaluated all models using the test-splits of the dataset they were fine-tuned with.}
Results are presented in Table~\ref{tab:nli} and show dominance of the \revision{double}-fine-tuned \XLM models. 
\revision{For reference, we also compare our results with previous research on English datasets -- in case of \FEVERNLI (and its Czech translation), our models achieved superiority over the NSMNs published in~\cite{nie2019combining} which scored an overall 69.5 macro-F1.
Our baselines for \CTKNLI and \FCZ (\textit{NearestP})  achieved and F-score of 76.9 and 83.2 percent, respectively.
This is comparable with the \cite{fever2018} baseline which scored 80.82\% \textit{accuracy} in a  sentence-level \textit{NearestP} setting on \FEN, using the Decomposable Attention model.
Interestingly, the \FCZ (\textit{NearestP}) experiment results are consistently higher than those of \FCZNLI{} -- we consider this estimate overly optimistic and attribute it to a possible partial information leakage discussed in Section~\ref{fcz-method} and the \FCZ noise examined in Section~\ref{fcz-validity}.}

\revision{
\begin{table}
\begin{center}
    \caption{\revision{F1 macro score (\%) comparison of our \BERT-like models fine-tuned for the \textit{NLI} task on \CTKNLI, \FCZ, and \FCZNLI datasets. Gold evidence was used as the NLI context for each claim. \FEVERNLI from~\cite{nie2019combining} listed for comparison with other research. B stands for \BERT architecture, R for \RoBERTa.
    }}\label{tab:nli}
\setlength\tabcolsep{5pt}
\begin{tabular*}{\textwidth}{llcc|cc|cc|cc}
    \toprule
    model & arch.         &\multicolumn{8}{c}{F-score by dataset (\%)}\\
    \midrule  
    \SMBERT &B-base	            &\multirow{10}{*}{\rotatebox{90}{\CTKNLI}}& 55.3	&\multirow{10}{*}{\rotatebox{90}{\FCZ (\textit{NearestP})}}& 82.8	&\multirow{10}{*}{\rotatebox{90}{\FCZNLI}}& 70.8	&\multirow{10}{*}{\rotatebox{90}{\FEVERNLI}}& 73.4\\
    \SlavicBERT & B-base	        && 60.6	        && 83.0	            && 70.3	&& --\\
    \CZERT  & B-base	            && 58.9	        && 80.7	            && 69.4	&& --\\
    \MBERT  & B-base	            && 58.1	        && 83.1	            && 70.9	&& 73.2\\
    \FERNETC & B-base	        && 63.8	        && 82.5	            && 71.9	&& --\\
    \FERNETN & R-base	        && 48.4	        && 81.3	            && 70.8	&& --\\
    \RobeCzech & R-base	    && 67.7	        && 82.2	            && 71.1	&& --\\
    \XLMSQUAD & R-base        && 61.2	        && 81.3	            && 69.8	&& 72.8\\
    \XLMSQUAD & R-large	    && \textbf{76.9}&& \textbf{83.2}	&& 72.2	&& \textbf{75.9}\\
    \XLMXNLI & R-large~~~ 	    && 74.3	&& 83.0	&& \textbf{73.7}	&& 75.4\\
    
    \bottomrule
\end{tabular*}
\end{center}
\end{table}
}
\begin{comment}
    \subsubsection{Machine-translated NLI corpora}
To address the \textit{overfitting} (Table~\ref{tab:overfitting}) issue in our future experiments, our team at \textsf{AIC} has internally obtained a machine-translated Czech localization for each of the corpora listed in~\ref{sec:nlicorp},  using the \textsf{Google Translate API}. The scheme is simpler than that from~\ref{sec:fevercs}, as their formats are closely resemblant to the \texttt{nli} \textsf{FCheck} export (Figure~\ref{list:fcheck-nli}), i.e., a set of plain text pairs and their labels.

In future, the Czech \textsf{SNLI}, \textsf{MultiNLI}, \textsf{ANLI} or \textsf{FEVERNLI} datasets could be either used directly to augment the \textsf{ČTK NLI train} dataset, or to construct a \textit{down-stream task} for NLI in Czech.
\end{comment}

%!TEX ROOT=../lrec2021.tex

\subsection{Full Pipeline Results}
\label{sec:full-pipeline}
% \begin{itemize}
    % \item \jd{Look at results from~\cite{fever2018b}}.
% \end{itemize}

Similarly to~\cite{fever2018}, we give baseline results for the full fact verification pipeline.
The pipeline is evaluated as follows: 1) given a mutated claim $m$ from the test set, $k$ evidence paragraphs (documents) $P = \{p_1, \ldots, p_k\}$ are selected using document retrieval models as described in Section~\ref{section:dr-baseline}.
Note that documents in $P$ are ordered by decreasing relevancy.
The paragraphs are subsequently fed to an NLI model of choice (see details below), and accuracy (for \FCZ and \FEN) or F1 macro score (for the unbalanced \CTK) are evaluated.
In case of supported and refuted claims, we analyze two cases: 1) for Score Evidence~(\textsf{SE}), $P$ must fully cover at least one gold evidence set, 2) for No Score Evidence~(\textsf{NSE}) no such condition applies.
No condition applies for \NEI{} claims as well.\footnote{Our approach corresponds to the \textsf{NearestP} strategy from~\cite{fever2018}. Similarly, \textsf{SE} and \textsf{NSE} correspond to \textsf{ScoreEv} and \textsf{NoScoreEv}.}

While our paragraph-oriented pipeline eliminates the need for sentence selection, we have to deal with the maximum input size of the NLI models (512 tokens in all cases), which gets easily exceeded for larger $k$.
Our approach is to iteratively partition $P$ into $n$ consecutive splits $S = \{s_1, \ldots, s_l\}$, where $l \leq n$.
Each split $s_i$ itself is a concatenation of successive documents $s_i = \{p_s, \ldots, p_e\}$, where $1\leq s \leq e\leq n$. 
A new split is created for any new paragraph that would cause input overflow. 
If any single tokenized evidence document is longer than the maximum input length, it gets represented by a single split and truncated\footnote{This is the only case in which truncation occurs.}.
Moreover, each split is limited to at most $k_s$  successive evidence documents ($k_s=2$ for \FCZ and \FEN, $k_s=3$ for \CTK), so the overall average input length is more akin to data used to train the NLI models.

In the prediction phase, all split documents $p_s, \ldots, p_e$ are concatenated, and, together with the claim $m$, fed to the NLI model getting predictions $y_s, \ldots, y_e$, where each $y_i = (y_i^{\SUP{}}, y_i^{\REF{}}, y_i^{\NEI{}})$ represents classification confidences.
Finally, the claim-level confidences are obtained as $y^c = \frac{1}{e-s}\sum_{i=0}^{e-s}\lambda^i y_i^c$ for $c \in \{\SUP{}, \REF{}, \NEI{}\}$.
This weighted average (we use $\lambda=\frac{1}{2}$ in all cases) assigns higher importance to the higher-ranked documents. 

\revision{
The results are presented in Table~\ref{table:full_pipeline}.
We evaluate \Anserini and \ColBERT DR models followed by overall best performing \XLMSQUAD NLI model (RoBERTa-large, described in Section~\ref{sec:nli-baseline}) for all datasets.
For both \FEVER-based datasets, \ColBERT document retrieval brings significantly the best results. 
For \CTK, \Anserini and \ColBERT perform similarly with \Anserini giving slightly better results overall which mimics the results of DR described in Section~\ref{section:dr-baseline}.
Note that the \textsf{SE} to \textsf{NSE} difference is more pronounced for \CTK, which can be explained by high redundancy of \CTK paragraphs w.r.t. \FCZ.
Comparing the results of the \FEN baseline to~\cite{fever2018} our models perform better in spite of discarding the sentence selection stage -- the best reported \FEN accuracies for 5 retrieved documents were $52.09\%$ fot \textsf{NSE} and $32.57\%$ for \textsf{SE}~\cite{fever2018}.
The improvement can be explained by more recent and sophisticated models used in our study (authors of~\cite{fever2018} used Decomposable Attention~\cite{parikh2016decomposable} at the NLI stage).
Note that \FEN full pipeline significantly outperforms \FCZ, which we explain by the information leakage and noise of the dataset used to train the \FCZ (\textit{NearestP}) model as discussed in the previous section.}
\begin{table}
    \begin{center}
    % \begin{minipage}{0.7\textwidth}
    \caption{\revision{Full pipeline results. Accuracy (\%) shown for \FCZ, and \FEN, F1 macro score (\%) for \CTK. Unlike \textsf{NSE} (No Score Evidence), \textsf{SE} (Score Evidence) demands correct evidence to be retrieved.}}\label{table:full_pipeline}
    \scalebox{0.85}{
    \begin{tabular}{llcccccccc}
    \toprule
    dataset & retrieval & \multicolumn{2}{c}{@1}  & \multicolumn{2}{c}{@5} & \multicolumn{2}{c}{@10} & \multicolumn{2}{c}{@20} \\
    & & \textsf{NSE} & \textsf{SE} & \textsf{NSE} & \textsf{SE} & \textsf{NSE} & \textsf{SE} & \textsf{NSE} & \textsf{SE}\\
    \midrule

    % first reviewed version
    % \multirow{3}{5em}{\textit{\FCZ rev1}}& \Anserini & 49.79&12.93 & 52.12&22.12 & 52.10&24.98 & 52.22&27.61\\
    % & \ColBERT{128} & 53.04&23.82 & 53.99&31.66 & 54.13&32.71 & 54.18&33.40\\
    % & \MBERT & \textit{53.93}&\textit{24.73} & \textit{55.40}&\textit{33.89} & \textit{55.40}&\textit{34.76} & \textbf{55.45}&\textbf{35.30}\\
    % \midrule
    % using old NLI models
    % \multirow{3}{5em}{\FCZ old NLI} &\DrQA&53.81&17.56&54.24&26.79&54.32&28.79&54.33&30.47\\
    % &\Anserini&51.67&15.10&53.37&23.11&53.51&26.15&53.68&28.86\\
    % &\ColBERT{128}&54.98&26.77&55.59&33.10&55.66&34.11&55.64&34.72\\
    % &\MBERT&\textit{56.02}&\textit{27.81}&\textit{56.59}&\textit{35.29}&\textit{56.65}&\textit{36.26}&\textbf{56.73}&\textbf{36.73}\\
    % \midrule

    % "/mnt/data/factcheck/nli_models/csfever/xlm-roberta-large-squad2_bs10_ep9_wr0.2"
    % removed DRQA (similar to Anserini) an M-BERT (missing for EnFEVER + failure for CTKFacts)
    % \multirow{3}{5em}{\FCZ} &\DrQA&52.89&16.04&52.69&19.11&52.64&19.24&51.23&17.39\\
    \multirow{3}{5em}{\FCZ} &\Anserini&49.13&14.17&50.59&18.69&50.65&19.89&50.67&20.65\\
    &\ColBERT{128}&\textbf{61.90}&\textit{25.67}&\textit{58.89}&\textit{29.81}&\textit{59.03}&\textit{30.31}&\textit{59.23}&\textbf{30.37}\\
    % &\MBERT&\textbf{63.58}&\textit{26.90}&\textit{61.16}&\textit{31.63}&\textit{61.20}&\textbf{32.00}&\textit{61.39}&\textit{31.85}\\
    \midrule

    % "/mnt/data/factcheck/nli_models/csfever/bert-base-multilingual-cased-sentence_bs8_ep20_wr0.3"
    % \multirow{3}{5em}{\FCZ BERT} &\DrQA&51.89&15.74&51.52&16.65&48.66&13.43&43.94&8.34\\
    % &\Anserini&48.32&13.88&49.04&16.63&49.03&16.80&47.42&14.38\\
    % &\ColBERT{128}&60.73&25.28&58.92&27.73&58.89&27.02&56.46&23.44\\
    % &\MBERT&\textbf{61.54}&\textit{26.22}&\textit{60.63}&\textbf{29.23}&\textit{60.45}&\textit{28.57}&\textit{57.90}&\textit{24.38}\\
    % \midrule

    % CTKFacts XNLI NLI models, commented out
    % \multirow{3}{5em}{\CTK XNLI} & \Anserini & \textit{60.86}&\textit{10.17} & 60.69&18.89 & 60.58&\textit{23.49} & 61.03&\textbf{26.91}\\
    % & \ColBERT{32} & 60.40&8.97 & \textit{61.24}&\textit{20.14} & \textbf{61.78}&22.94 & \textit{61.74}&26.08 \\
    % \midrule

    % CTKFacts SQUAD2 NLI models
    \multirow{3}{5em}{\CTK} & \Anserini &\textit{60.06}&\textit{11.24}&\textit{58.35}&\textit{19.97}&\textbf{59.69}&\textit{24.79}&\textit{59.50}&\textbf{26.71}\\
    & \ColBERT{32} &59.94&10.68&56.97&19.28&57.23&22.74&56.39&25.51 \\
    \midrule

    % EnFEVER SQUAD2 NLI models, DRQA and XNLI models commented out
    % \multirow{3}{5em}{\FEN} & \DrQA& 60.03&17.29& 60.60&29.31& 60.28&33.81& 60.21&37.42\\
    % & \DrQA XNLI& 61.19&16.88& 61.21&28.87& 61.44&33.60& 61.46&37.21\\
    \multirow{3}{5em}{\FEN}& \Anserini& 66.47&22.44 & 65.72&34.10 & 65.59&35.97 & 65.37&35.39\\
    % & \Anserini XNLI& 66.87&22.44 & 64.05&33.93 & 64.47&36.04 & 64.51&35.75\\
    & \ColBERT{128}&\textbf{68.19}&\textit{33.83}&\textit{66.21}&\textit{46.15}&\textit{65.73}&\textit{48.20}&\textit{65.43}&\textbf{49.16}\\
    % & \ColBERT{128} XNLI &\textit{76.46}&\textit{50.54}&\textit{79.81}&\textit{68.26}&\textbf{79.97}&\textit{71.89}&\textit{79.84}&\textbf{73.67}\\
    \bottomrule
    \end{tabular}
    }
    % \end{minipage}
    \end{center}
\end{table}

%\input{src/discussion}
\begin{comment}
\input{src/licensing}
\end{comment}
\section{Conclusion}\label{sec:conclusion}

With this \revision{article}, we \revision{examined two major ways to acquire Czech data for automated fact-checking.}

Firstly, we localized the \FEN dataset, using a document alignment between Czech and English \Wikipedia abstracts extracted from the \textit{interlingual links}.
We obtain and publish the \FCZ dataset of 127k machine-translated claims with evidence enclosed within the Czech \Wikipedia dump.
We then validate our alignment scheme and measure a 66\% precision using hand annotations over a 1\% sample of obtained data.
Therefore, we recommend the data for models less sensitive to noise and
\revision{we proceed to utilize \FCZ for training non-critical retrieval models for our annotation experiments and for recall estimation of our baseline models.
Furthermore, we publish a \FCZNLI dataset of 228k context-query pairs directly translated from English to Czech that bypass its issue with noise for the subroutine task of Natural Language Inference.}

Secondly, we executed a series of human annotation runs with 163 students of journalism to acquire a novel dataset in Czech.
As opposed to similar annotations that extracted claims and evidence from \Wikipedia~\cite{fever2018,norregaard2021danfever,aly2021feverous}, we annotated our dataset on top of a CTK corpus extracted from a news agency archive to explore this different relevant language form. 
We collected a raw dataset of 3,116 labeled claims, 57\% of which have at least two independent cross-annotations.
From these, we calculate Krippendorff's alpha to be 56.42\% and 4-ways Fleiss' $\kappa$ to be 63\%.
We proceed with manual and human-and-model-in-the-loop annotation cleaning to remove conflicting and malformed annotations, arriving at the thoroughly cleaned \CTK dataset of 3,097 claims and their veracity annotations complemented with evidence from the CTK corpus.
We release its version for NLI called \CTKNLI to maintain corpus trade secrecy.

Finally, we use our datasets to train baseline models for the full fact-checking pipeline composed of Document Retrieval and Natural Language Inference tasks.
% At five retrieved paragraphs in a Score Evidence setting, where at least one gold set of evidence must be covered by the DR predictions, our best performing models scored 33.89\% overall accuracy on \FCZ and 20.14\% on \CTK.
% In a No Score Evidence setting, where the gold evidence requirement is removed, we scored 55.40\% and 61.24\% on \FCZ and \CTK, respectively.
% We claim the results to be testifying to the viability of the task in our setting and encouraging for further research on our data.

\subsection{Future work}
\begin{itemize}
    \item The fact-checking pipeline is to be augmented by the \textit{check-worthiness estimation}~\cite{nakov2021automated}, that is, a model that classifies which sentences of a given text in Czech are appropriate for the fact verification.
    We are currently working on models that detect claims within the Czech Twitter, and a strong predictor for this task would also strengthen our annotation scheme from Section~\ref{sec:annotation} that currently relies on hand-picked check-worthy documents.
    \item While the \SUP, \REF{} and \NEI{} classes  offer a finer classification w.r.t. evidence than binary \texttt{true}/\texttt{false}, it is a good convention of fact-checking services to use additional labels such as \texttt{MISINTERPRETED}, that could be integrated into the common automated fact verification scheme if well formalised.
    \item The claim extraction schemes like that from~\cite{fever2018} or Section~\ref{sec:annotation} do not necessarily produce organic claims capturing the real-world complexity of fact-checking.
    For example, just the \FEN \train set contains hundreds of claims of form \enquote{X is a person.}.
    This problem does not have a trivial solution, but we suggest integrating real-world claims sources, such as Twitter, into the annotation scheme.
    \item While the \FEVER localization scheme from Section~\ref{fcz-method} yielded a rather noisy dataset, its size and document precision encourage
    deployment of a model-based cleaning scheme like that from~\cite{Jeatrakul} to further refine its results.
    \revision{I. e., a well performing NLI model could do well in pruning the invalid datapoints of \FCZ without further annotations.}
\end{itemize}

\backmatter

% \bmhead{Supplementary information}

% If your article has accompanying supplementary file/s please state so here. 

\bmhead{Acknowledgments}
This article was produced with the support of the Technology Agency of the Czech Republic under the ÉTA Programme, project TL02000288. The access to the computational infrastructure of the OP VVV funded project CZ.02.1.01/0.0/0.0/16\_019/0000765 ``Research Center for Informatics'' is also gratefully acknowledged.
We would like to thank all annotators as well as other members of our research group, namely: Barbora D\v{e}dkov\'{a}, Alexandr Ga\v{z}o, Jan Petrov, and Michal Pitr.

\begin{comment}
\section*{Declarations}

\begin{itemize}
\item Funding
\item \textbf{Conflict of interest} The author declares that he has no conflict of interest.
% \item Ethics approval 
% \item Consent to participate
% \item Consent for publication
\item Availability of data and materials
\item Code availability 
% \item Authors' contributions
\end{itemize}
\end{comment}


\bibliography{lrec2021}
\pagebreak

\begin{appendices}
\section{Annotation platform}\label{sec:annotation_platform}
\label{appendix:annotations}
\subsection{Claim Extraction}

\begin{figure}
    \includegraphics[width=\textwidth]{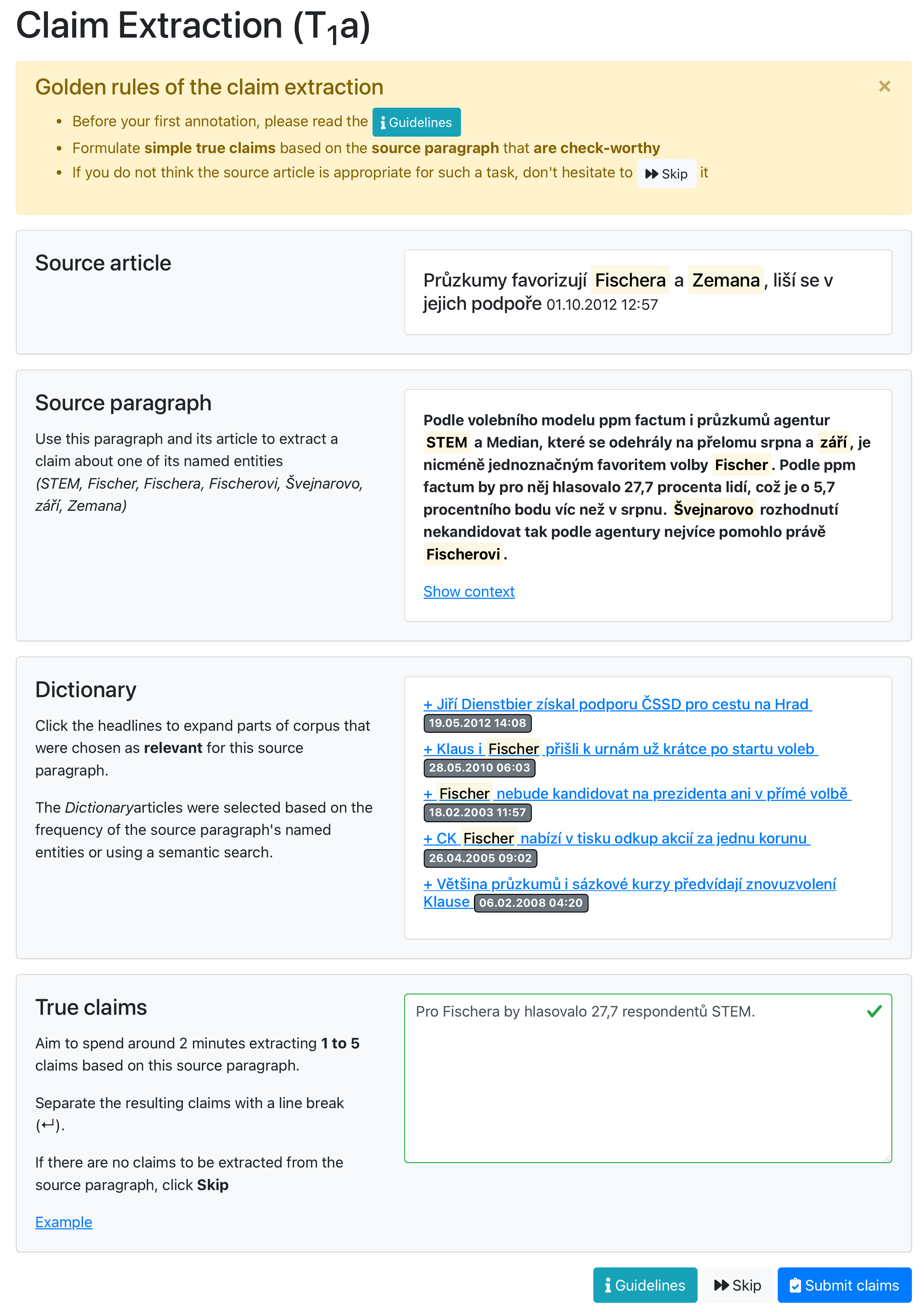}
    \label{fig:extraction}
    \caption[Claim extraction interface of \textsf{FCheck} platform]{Claim extraction interface of the \textsf{FCheck} platform. Captions were translated to English, corpus data kept in Czech.}
\end{figure} % Extraction

Figure~\ref{fig:annotation} shows the \textit{Claim Extraction} (\ToneA) interface.

The layout is inspired by the work of~\cite{fever2018} and, by default, hides as much of the clutter away from the user as possible.
Except for the article heading, timestamp, and source paragraph, all supporting information is collapsed and only rendered on user demand. 

An annotator reads the source article and, if it lacks a piece of information he/she wants to extract, looks for it in the expanded article or corpus entry.
The user is encouraged to {Skip} any source paragraph that is hard to extract -- the purpose of \textit{Source Document Preselection} (\Tzero) was to reduce these skips as much as possible.

Throughout the platform, we have ultimately decided not to display any \textit{stopwatch}-like interface not to stress out the user.
We have measured that, excluding the outliers ($\leq10s$, typically the {Skip}ped annotations, and $\geq 600s$, typically a browser tab left unattended), average time spent on this task is \textbf{2 minutes 16 seconds} and the median is \textbf{1 minute 16 seconds}.

\subsection{Claim Mutation}
\begin{figure}
    \includegraphics[width=\textwidth]{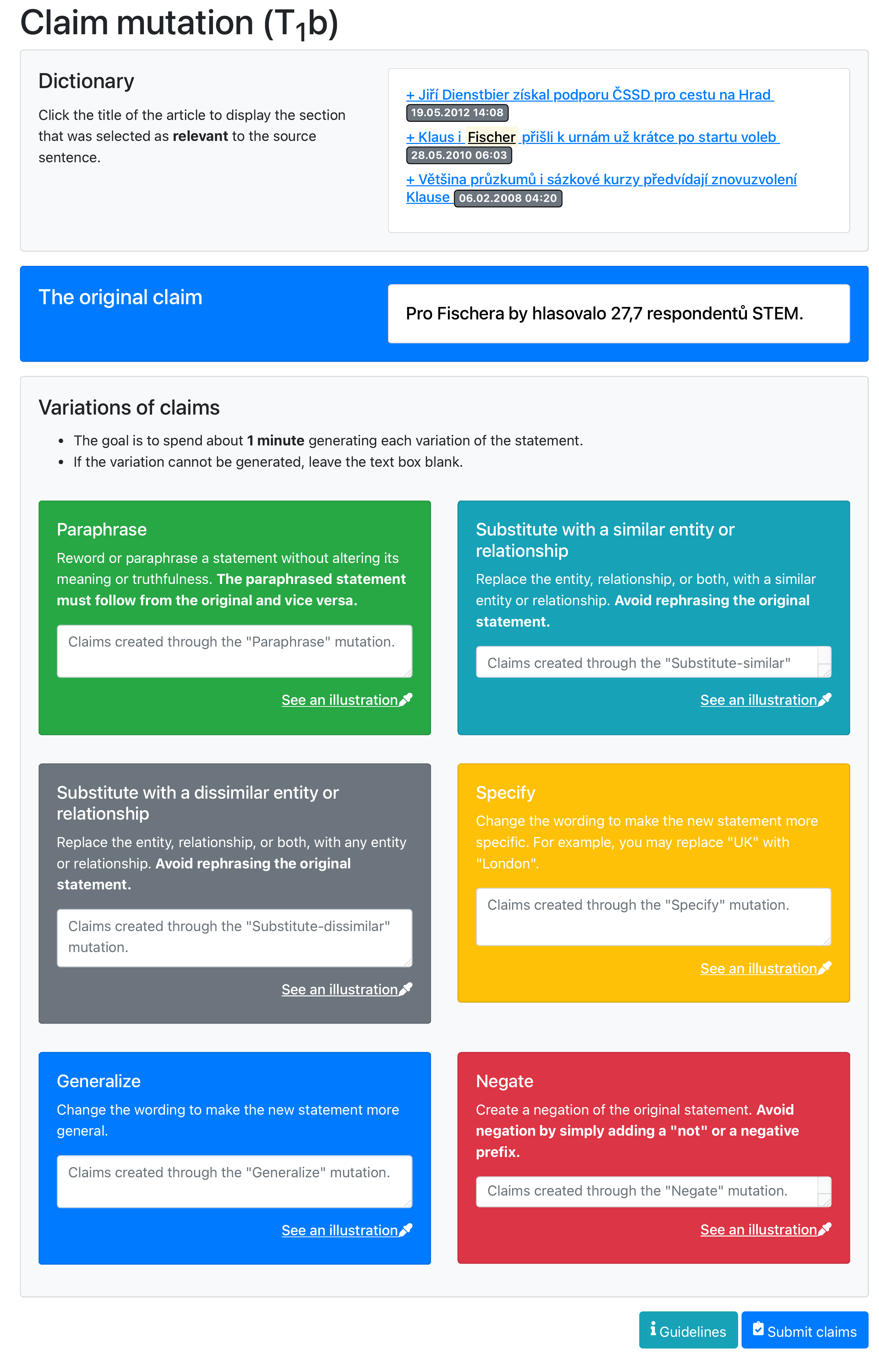}
   
   \caption[Claim mutation interface of \textsf{FCheck} platform]{Claim mutation interface. Captions were translated to English, corpus data kept in Czech.}    
   \label{fig:mutation}
\end{figure}

Mutation types follow those of the \textsf{\FEVER Annotation Platform} and are distinguished by loud colors, to avoid mismatches.

Excluding the outliers, the overall average time spent generating a batch of mutations was \textbf{3m 35s} (median \textbf{3m 15s}) with an average of \textbf{3.98} mutations generated per claim.

\subsection{Claim Veracity Labeling}
\begin{figure}
    \includegraphics[width=\textwidth]{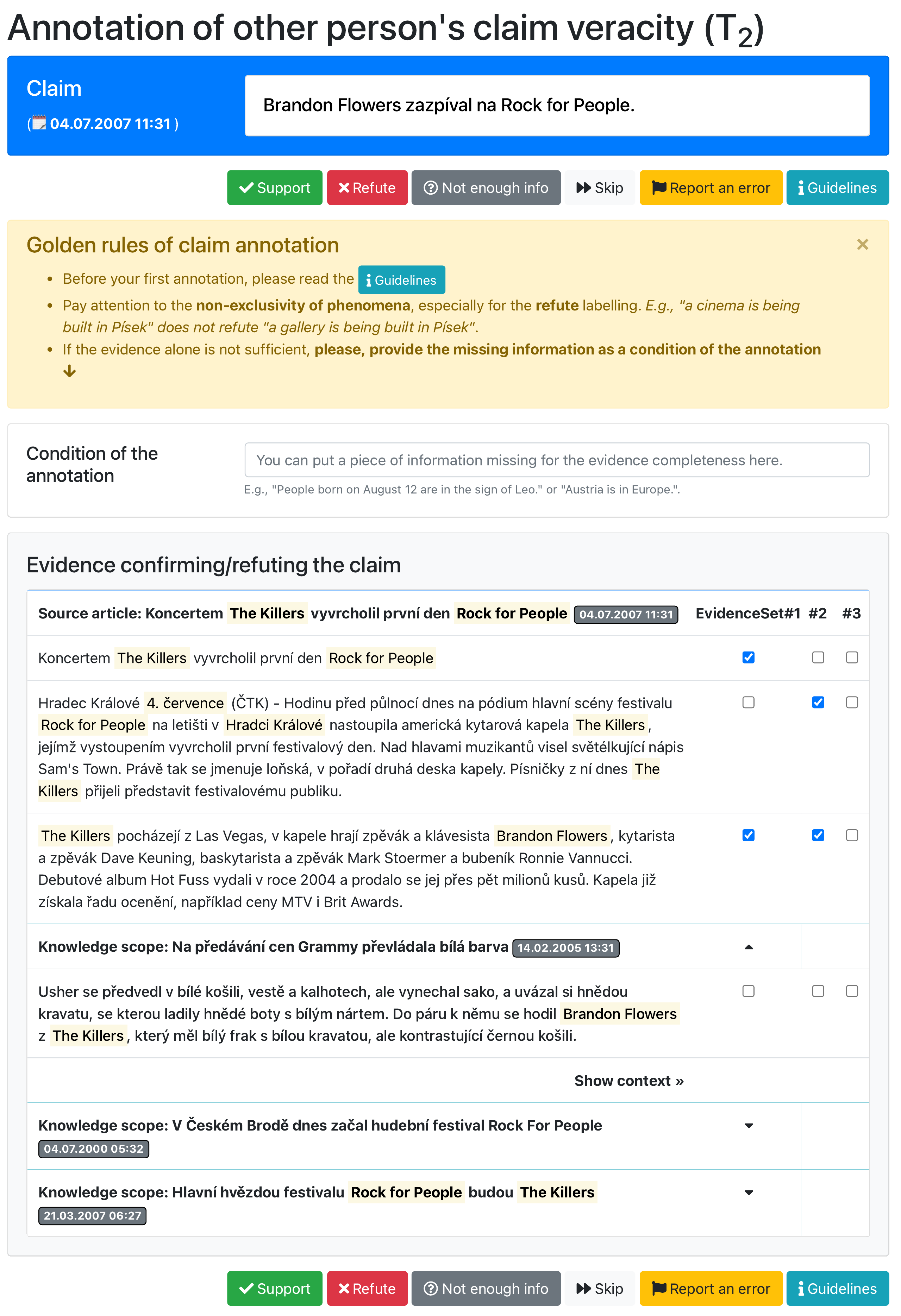}
   
   \caption[Claim labelling interface of \textsf{FCheck} platform]{Claim labelling interface. Captions were translated to English, corpus data kept in Czech.}    
   \label{fig:annotation}
\end{figure}

\label{sec:ui-labeling}
In Figure~\ref{fig:annotation} we show the most complex interface of our platform -- the \Ttwo{}: \textbf{Claim Annotation} form. Full instructions took about 5 minutes to read and comprehend.

The input of multiple evidence sets works as follows: each column of checkboxes in~\ref{fig:annotation} stands for a single evidence set, every paragraph from the union of knowledge belongs to this set iff its checkbox in the corresponding column is checked. Offered articles and paragraphs are collapsible, empty evidence set is omitted.

On average, the labelling task took \textbf{65 seconds}, with a median of \textbf{40s}.
An average \texttt{SUPPORTS}/\texttt{REFUTES} annotation was submitted along with \textbf{1.29} different evidence sets, 95\% of which were composed of a single paragraph.

\end{appendices}

\end{document}